\PassOptionsToPackage{table}{xcolor}
\documentclass[10pt, logo, copyright]{nv}

\usepackage{amsmath,amsfonts,bm}

\def\eqref#1{equation~\ref{#1}}

\def\1{\bm{1}}

\DeclareMathAlphabet{\mathsfit}{\encodingdefault}{\sfdefault}{m}{sl}
\SetMathAlphabet{\mathsfit}{bold}{\encodingdefault}{\sfdefault}{bx}{n}

\providecommand{\methodname}{Physis-Lang}

\newcommand{\hide}[1]{}

\usepackage{hyperref}
\usepackage{url}
\usepackage{booktabs}
\usepackage{float}
\usepackage{array}
\usepackage{colortbl}
\definecolor{tablegray}{gray}{0.92}
\usepackage{graphicx}
\usepackage{subcaption}
\usepackage{placeins}
\usepackage[numbers]{natbib}
\def\tablecite#1{[\citenum{#1}]}

\usepackage{booktabs} 
\usepackage{caption}  
\usepackage{pifont}

\usepackage{amssymb}
\usepackage{subcaption}
\usepackage[table]{xcolor}
\usepackage{xcolor}   
\usepackage{multirow}
\usepackage{makecell} 
\usepackage{url}
\usepackage{booktabs}
\usepackage{graphicx}
\usepackage{animate}
\usepackage{xspace}
\usepackage{amsmath,amsfonts,bm}
\usepackage{enumitem}
\usepackage{cite}
\usepackage[ruled]{algorithm2e}

\usepackage{cleveref}
\crefname{section}{Sec.}{Secs.}
\crefname{table}{Tab.}{Tabs.}
\crefname{figure}{Fig.}{Figs.}
\crefname{algorithm}{Alg.}{Algs.}

\definecolor{cvprblue}{rgb}{0.21,0.49,0.74}
\definecolor{iccvblue}{rgb}{0.21,0.49,0.74}
\definecolor{mitblue}{rgb}{0.88,0.95,0.96}
\definecolor{gold}{rgb}{0.75,0.6,0.12}
\colorlet{shadecolor}{gray!40}
\definecolor{mydarkred}{rgb}{0.8,0.02,0.02}

\def\tablecite#1{[\citenum{#1}]}
\newcolumntype{g}{>{\columncolor{mitblue}}c}
\newcolumntype{f}{>{\columncolor{mitblue}}l}
\newcolumntype{h}{>{\columncolor{mitblue}}r}
\newcolumntype{i}{>{\columncolor{gray}}c}

\usepackage[most]{tcolorbox}

\title{
  Physis-Lang: Self-Evolving Language as a Physical Representation for Video World Model}

\author{
  {\small\bfseries
    Liming Lu$^{\dagger 2}$,
    Xianzheng Ma$^{\dagger 3}$,
    Wenkun He$^{\dagger 2}$,
    Guanqi Zhan$^{\dagger *1}$
  } \\
  {\small\bfseries
    Yilin Zhao$^{1}$,
    Junyu Chen$^{1}$,
    Mengyao Xu$^{1}$,
    Jiaojiao Fan$^{1}$,
    Wenhang Ge$^{1}$,
    Yuchao Gu$^{1}$,
    Yunze Liu$^{1}$
  } \\
  {\small\bfseries
    Boyi Li$^{1}$,
    Zhen Dong$^{1}$,
    Victor Prisacariu$^{3}$,
    Ming-Yu Liu$^{1}$,
    Song Han$^{1}$,
    Han Cai$^{*1}$
  } \\~\\
  $^{1}$NVIDIA ~ $^{2}$MIT ~ $^{3}$University of Oxford \\
  $^\dagger$ Equal Contribution, *corresponding \\
}
\correspondingauthor{Guanqi Zhan (gzhan@nvidia.com), Han Cai (hcai@nvidia.com)}

\begin{abstract}
Video world models are expected to predict how the physical world evolves, yet they often produce visually plausible videos that violate basic physical principles. Existing approaches commonly assume that natural language is insufficient to represent the physical knowledge required for reliable generation, and therefore introduce additional visual, latent, numerical, or planning-based signals. 
We revisit this assumption and introduce \emph{\methodname}, a self-evolving framework that treats physical language as a shared and optimizable representation across data curation, model training, and video generation.
\methodname{} represents physical processes through language that describes their relevant entities, causes, interactions, governing principles, temporal evolution, and effects. To improve this representation, we construct \emph{PhysCapBench}, which decomposes physical processes into atomic assertions and evaluates captions using recall and precision. An agentic loop iteratively analyzes assertion-level errors and refines the instruction used to produce physical captions. 
\methodname{} further converts model deficiencies into textual descriptions and uses language-guided retrieval to identify visually diverse videos that cover missing physical processes. 
Experiments on four widely used physical video benchmarks with Wan and Cosmos backbones demonstrate consistent improvements in physical plausibility. 
Notably, starting from open-source Cosmos3-Nano backbones, our Physis-Lang-enhanced models surpass the leading proprietary Veo~3.1 model.

\end{abstract}

\begin{document}
\maketitle

\begin{abstract}
Video world models are expected to predict how the physical world evolves, yet they often produce visually plausible videos that violate basic physical principles. Existing approaches commonly assume that natural language is insufficient to represent the physical knowledge required for reliable generation, and therefore introduce additional visual, latent, numerical, or planning-based signals. 
We revisit this assumption and introduce \emph{\methodname}, a self-evolving framework that treats physical language as a shared and optimizable representation across data curation, model training, and video generation.
\methodname{} represents physical processes through language that describes their relevant entities, causes, interactions, governing principles, temporal evolution, and effects. To improve this representation, we construct \emph{PhysCapBench}, which decomposes physical processes into atomic assertions and evaluates captions using recall and precision. An agentic loop iteratively analyzes assertion-level errors and refines the instruction used to produce physical captions. 
\methodname{} further converts model deficiencies into textual descriptions and uses language-guided retrieval to identify visually diverse videos that cover missing physical processes. 
Experiments on four widely used physical video benchmarks with Wan and Cosmos backbones demonstrate consistent improvements in physical plausibility. 
Notably, starting from open-source Cosmos3-Nano backbones, our Physis-Lang-enhanced models surpass the leading proprietary Veo~3.1 model.

\end{abstract}    
\begin{figure}[H]
	\centering
	\vspace{-0.5cm}
    \includegraphics[width=\linewidth]{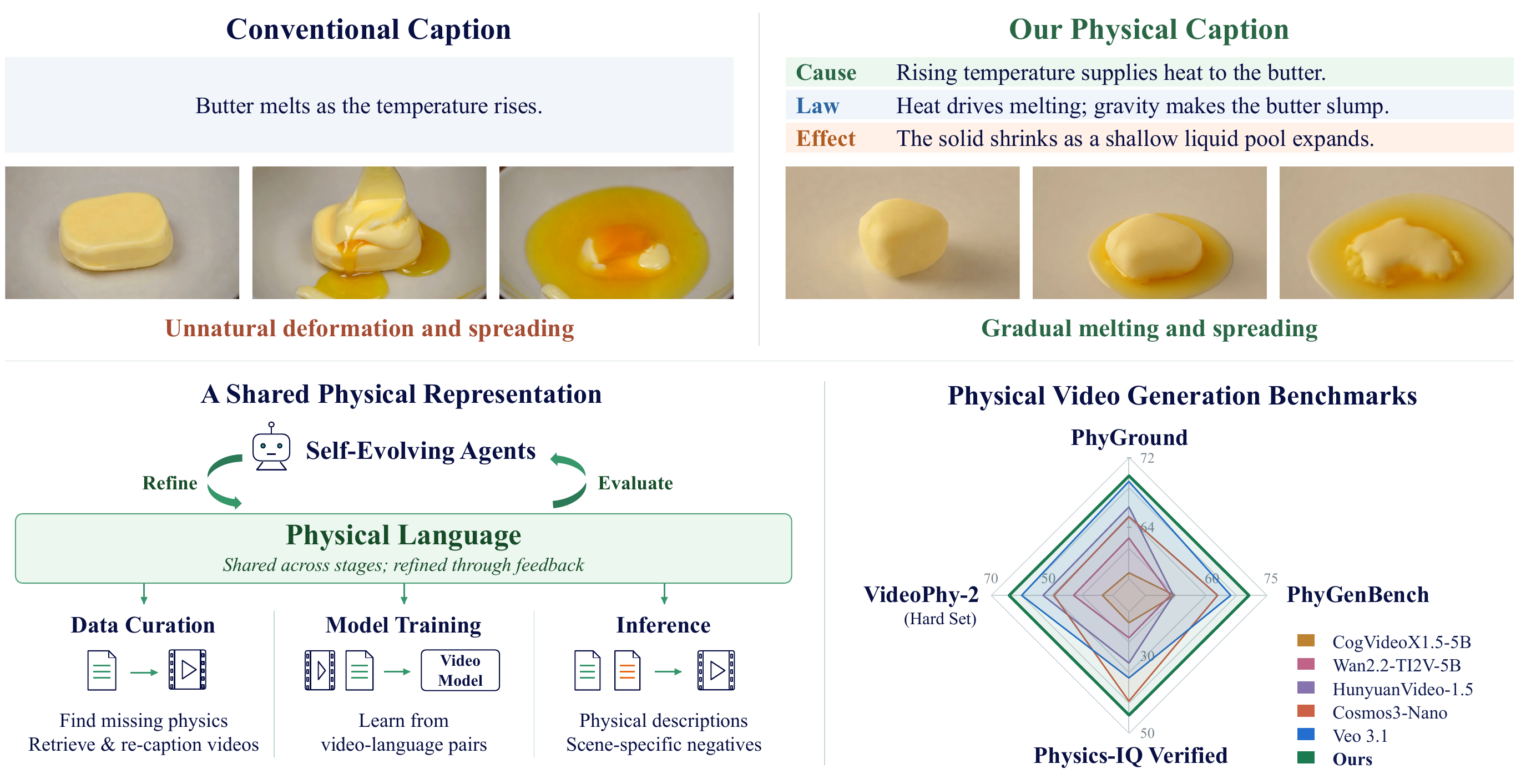}
    \vspace{-0.6cm}
	\caption{\textbf{Overview of \methodname}. Existing approaches treat language mainly as a conditioning interface and introduce physical knowledge through separate visual, latent, numerical, or planning-based signals. \methodname{} instead treats physical language as a shared and optimizable representation: an agentic loop evaluates and refines physical descriptions, language-guided retrieval expands the training data toward missing physical processes, and the evolved language supervises training and guides inference toward physically plausible video generation.}
	\label{fig:phylang_intro}
\end{figure}
\section{Introduction}
\label{sec:intro}

Video world models are expected not only to synthesize visually appealing videos, but also to predict how the physical world evolves.
This capability is fundamental to embodied intelligence, robotics, autonomous systems, and interactive simulation, where a model must anticipate the consequences of actions rather than merely render plausible-looking frames.
However, conventional video--text training primarily optimizes visual fidelity and semantic alignment, without explicitly representing the physical knowledge underlying observed dynamics; as a result, current video generation models still frequently violate basic physical principles: objects deform or disappear unexpectedly, collisions produce implausible outcomes, fluids move unnaturally, and causal events occur in the wrong order.
Recent benchmarks have further shown that perceptual realism does not necessarily imply physical understanding~\citep{phygenbench2024,videophy2_2025,physicsiq2025,paibench2025}.
Improving the \emph{physical plausibility} of video world models therefore remains a central challenge on the path toward general-purpose world simulation.

Existing efforts improve the physical plausibility of video generation by complementing the standard language interface with additional sources of physical information. They enrich video--text data with physics-focused videos or physical annotations~\citep{pisa2025,physinone2026}, introduce auxiliary signals about geometry, motion, or physical correctness during training~\citep{phantom2026,newtongen2026}, and provide structured plans, retrieved examples, or external judgments during generation~\citep{reasoningimplausibility2025,causalmotion2026}. Despite their different implementations, these approaches reflect a common design choice: natural language is primarily used to specify semantic content, while more detailed physical information is conveyed through complementary visual, geometric, latent, numerical, or human-designed signals. 
These complementary signals can provide useful supervision for aspects of physical processes that language alone may not fully capture. 
Yet an important possibility remains relatively unexplored: \textit{can language itself, when made more explicit and systematically optimized, serve as a unified representation of physical knowledge across the video-generation pipeline?}

Motivated by this question, we hypothesize that the limited role of language in current video models may stem from language not yet being fully exploited as a representation of physical processes, rather than from an intrinsic limitation of language itself. 
Based on this view, we introduce \emph{\methodname}, a self-evolving agentic framework that constructs, evaluates, and refines physical language as a shared representation, as illustrated in Fig.~\ref{fig:phylang_intro}. 
At the caption level, \methodname{} goes beyond describing visible actions and makes the relevant physical entities, causal interactions, governing principles, and resulting effects explicit. The self-evolving representation is then reused throughout the pipeline: it guides data curation by identifying missing physical domains and selecting relevant videos, provides supervision for training on video--language pairs, and guides inference through scene-specific physical descriptions. In this way, the three components shown at the bottom of Fig.~\ref{fig:phylang_intro} are connected by the shared physical language at its center, enabling the system to organize, evaluate, and transfer physical knowledge across data curation, training, and inference.

To build this shared representation, we first instantiate it at the caption level through an agentic caption-evolution loop grounded by our \emph{physics-aware critic} and \emph{PhysCapBench}. Given a fixed captioning model and an initial instruction, the model generates candidate physical descriptions for videos in our development set. For each video, we provides human-curated reference assertions organized around \emph{cause}, \emph{physical law}, and \emph{effect}. Using these references, our critic evaluates each candidate description along two dimensions: its coverage of the relevant physical knowledge and the faithfulness of its individual claims to the visual evidence. An evolution agent uses this diagnostic feedback to identify omissions, unsupported claims, and vague causal relations, and then revises the captioning instruction without updating the captioner's weights. Each instruction is then evaluated on our PhysCapBench within the same loop, which continues until benchmark performance saturates. The best-performing instruction is then used to re-caption the training corpus.
Thus, our physics-aware critic provides the loop feedback and PhysCapBench provides stopping criterion for language evolution, while \methodname{} turns that feedback into improved physical descriptions for video-model training and inference.

The self-evolving physical language also enables \methodname{} to expand the training distribution toward physical domains that the target video model handles poorly. It first summarizes the model's failures into a category-level deficiency profile. These deficient domains are then matched against physics tags associated with videos in a large gallery. Unlike retrieval based solely on visual similarity, this language-guided matching selects videos according to their physical content rather than their appearance, allowing the data engine to target missing physical domains and collect visually diverse examples relevant to the same deficiency. The selected videos are re-captioned with the self-evolving guidelines and added to the training corpus, directing additional supervision toward the model's weaknesses. Together with positive physical reasoning and scene-specific negative descriptions at inference time, this language-guided data engine connects targeted data curation, model training, and generation through a shared physical-language representation.

Our main contributions are summarized as follows:
\begin{itemize}
    \item We introduce a new perspective that treats \textbf{language as a physical world representation}. 
We show that language, when explicitly structured and self-evolving, can serve as a representation
for carrying, refining, and applying physical knowledge in video world models.

    \item We develop a \textbf{self-evolving agent system} that optimizes physical-language guidelines using a physics-aware critic. We introduce \textbf{PhysCapBench} to evaluate physical coverage and claim faithfulness and to guide iterative refinement.

    \item We build a \textbf{language-guided data engine} that uses physics-domain tags and text-level matching to select training videos relevant to model deficiencies, then enriches them with self-evolving physical captions.

    \item Extensive experiments on four physical video benchmarks demonstrate significant and consistent improvements on both Wan and Cosmos backbones. Our resulting models also outperform Veo~3.1 under the same evaluation protocols, validating language as a practical representation for improving video world models.
\end{itemize}

\section{Related Work}
\label{sec:related}

\vspace{2pt} \noindent \textbf{Physics-Aware Video Generation.} 
Recent efforts to enhance the physical plausibility of video generative models can be categorized by the source of the physical priors they exploit. The largest line of work draws physical priors from pretrained vision-language or foundation models, utilizing them to distill implicit physical dynamics into auxiliary branches~\citep{moalign2025, physvideogenerator2026, phantom2026, mmphysvideo2026, ditmem2025, tsa2026, prophy2026}, retrieve physically grounded reference motions~\citep{ragme2025, motionrag2025, physrag2026}, verify or critique generated contents~\citep{reasoningimplausibility2025, pris2026, seekingphysics2026, wmreward2026, latsearch2026, diffphy2025,xue2025phyt2v,liu2025bootstrappingphysicsgroundedvideogeneration}, or plan over intermediate representations before rendering~\citep{physvid2026, newton2026, chainofevent2026, vchain2025, shang2026phizero}. A second line treats physical plausibility as an objective to be optimized via reinforcement learning or preference optimization, aligning the generator with reward or preference signals derived from physics-aware judges or contrastive trajectories~\citep{physhpo2025, physcorr2025, phyworld2026, diffusionapo2026, psdpo2026, dynamicsboost2026, phyprompt2026, diffusiondrf2026, direct2026}. A third line of works injects explicit graphics- or simulator-derived structures like physical equations, trajectories, or executable simulation code into the generation pipeline~\citep{newtongen2026, stance2025, phdreamer2026, envisioningfuture2026, longtermmotion2026, causalmotion2026, tgt2025, motionforcing2026, moregen2026, videococo2026}. Some works also curate large-scale physics-annotated datasets~\citep{physinone2026, pisa2025, phyparam2026}.

Despite these advances, most existing approaches rely on auxiliary mechanisms, such as reward models, intermediate features, or retrieved samples, to incorporate physical knowledge.
Simulation-based approaches, on the other hand, are constrained by high computational costs and limited generalizability. In contrast, we posit that language provides a compact, explicit, and transferable representation of physical knowledge. Building on this view, we introduce an agentic self-evolution prompting scheme that iteratively refines the language representation to better capture physical priors. Different from the views of many recent works, we prove that \textbf{\textit{language alone}} is very powerful to lead to physically plausible video generation if we utilize it properly.

\vspace{2pt} \noindent \textbf{Agentic Self-Evolution.} 
Self-evolution has emerged as a distinct paradigm for improving agents, which has been applied to a range of tasks, including web navigation~\citep{agentq2024, webevolver2025, bagel2024, openwebvoyager2024}, mathematical and code reasoning~\citep{rzero2025, textgrad2024, gepa2026, dynamiccheatsheet2025, ace2026, godelagent2025, pace2026}, general instruction-following and question answering~\citep{selfinstruct2023, wizardlm2023}, as well as interactive tool use~\citep{agentevolver2025, evods2026, worldknowledgeexploration2026}. Across these tasks, an agent's prompts, policy, memory, or tools are iteratively refined. Representative methods involve revising subsequent attempts through verbal self-critique of past failures~\citep{reflexion2023, selfrag2023, webcot2025}, using the model itself as a judge to construct preference data for iterative alignment training~\citep{restreact2023, selfrewarding2024}, or training agents via multi-turn reinforcement learning within an interactive environment~\citep{ragen2025, evotrainer2026, seal2026}.  In contrast, we are the first to apply agentic self-evolution to physics reasoning that ultimately enhances the physical realism of video generation tasks. We establish a physics caption benchmark to evaluate the agents, which allows the upsampled prompts to converge toward a format that is optimally suited for generating physically realistic videos.
\section{Method}
\label{sec:method}

\subsection{Self-Evolving Agents to Improve Physics Caption} 
\label{sec:prompt_description}
\begin{figure}[t]
    \centering
    \includegraphics[width=\linewidth]{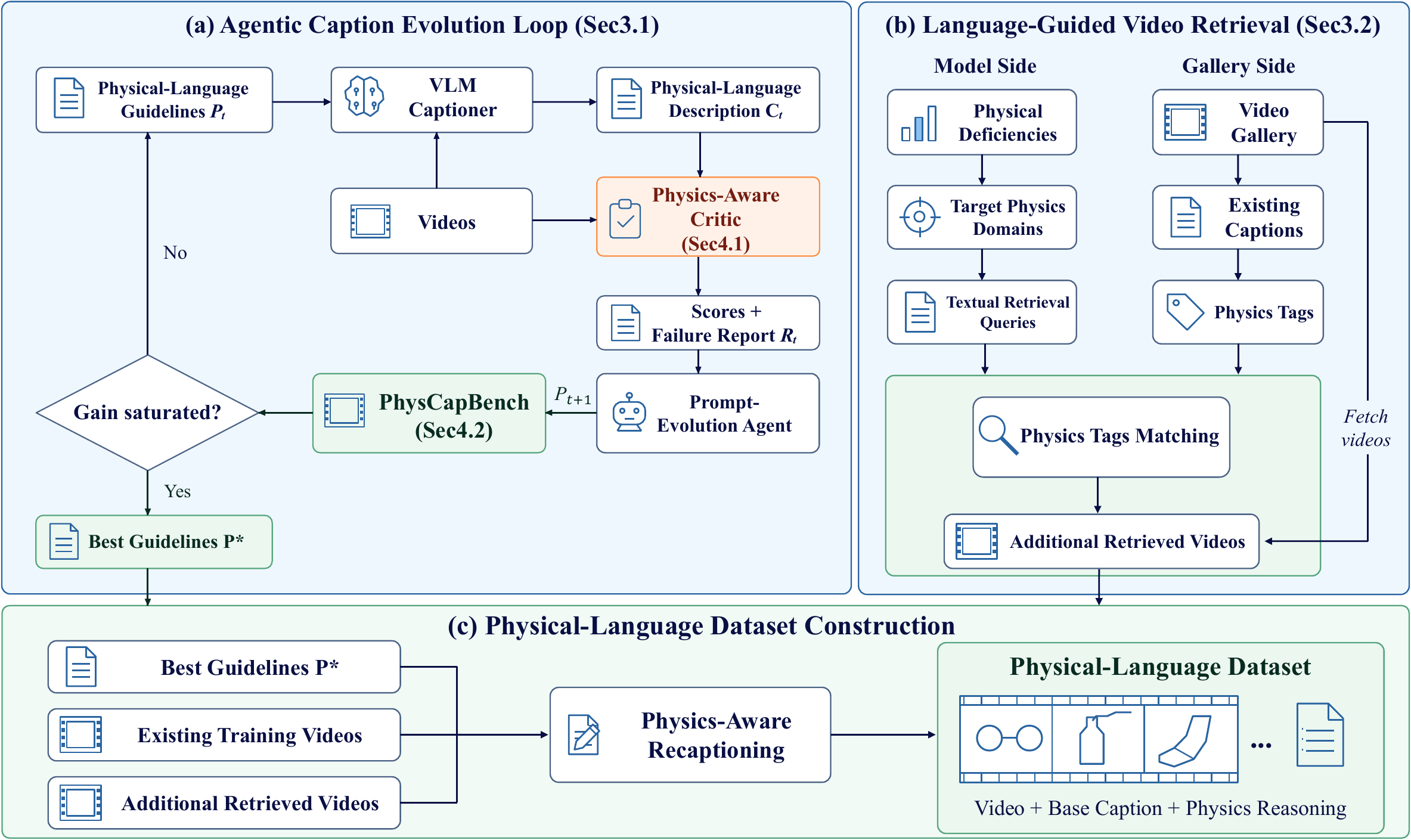}
    \caption{\textbf{Overview of our method}. Physical processes are represented through natural language, including a base caption, explicit physics reasoning, and scene-adaptive negative physics descriptions. A PhysCapBench-driven agentic loop evaluates and refines this physical language, which is then used to expand the training data, supervise video-model training, and guide inference toward physically plausible generation.}
    \label{fig:phylang_framework}
\end{figure}

Fig.~\ref{fig:phylang_framework}(a) demonstrates the overview of our agentic caption evolution loop. Conventional captions often describe scene semantics without explaining physical causes and consequences. We make these mechanisms explicit in language so that video--text training can provide richer supervision for physical dynamics. To this end, we augment base caption with a natural-language \textbf{\textit{physics\_reasoning}} field describing physical details, dynamics, and causal relations. The same representation supports training supervision and image/text-conditioned prompt expansion at inference.
Because captioning instructions can omit key details or induce unsupported reasoning, we use a self-evolving agent to iteratively refine the instruction while keeping the captioning model fixed.

To enable systematic prompt evolution, we first develop a \textbf{physics-aware critic} that evaluates the correctness and completeness of physical descriptions. The detailed protocol is introduced in Sec.~\ref{sec:critic—only}. Based on this critic, we further introduce \textbf{PhysCapBench}, a held-out benchmark of 246 physical videos for evaluating physics-aware video captioning. It serves both as a standalone benchmark and as the validation set for prompt evolution.

We then construct an agentic self-evolution loop over a fixed 20-video development set. At iteration $t$, a fixed captioner uses prompt $P_t$ to generate captions $\mathcal{C}_t$. The critic produces a score and diagnostic report $R_t$, which summarizes errors and guides an evolution agent to revise the prompt:
$P_t \rightarrow \mathcal{C}_t \rightarrow R_t \rightarrow P_{t+1}$.
After each iteration, the updated prompt is evaluated on PhysCapBench. We continue the loop while benchmark performance improves and stop once the gain saturates, selecting the best-performing prompt as the final physics-captioning instruction. Detailed implementation is provided in App.~\ref{app:loop_detail}.

During training, we apply the final captioner to re-caption the collected videos with physics-rich descriptions. During inference, we construct a corresponding upsampler that expands short text prompt and an input image (optional) into a detailed physics-aware prompt, providing the video generator with more explicit physical guidance and improving the physical realism of the generated video.
We further derive a scene-specific \textbf{\textit{physics\_negative\_prompt}} describing likely physically implausible evolutions, which serves as negative conditioning to suppress physical inconsistencies and improve generation realism at inference time.

\subsection{Language-Guided Video Retrieval}
\label{sec:video_retrieval}

Beyond improving the linguistic representation of physical processes, we further leverage language as a key modality for targeted data expansion, using it to identify and collect training examples that address the physical deficiencies of pretrained video generative models. Fig.~\ref{fig:phylang_framework}(b) illustrates the framework of this data collecting pipeline.

We first build a GPT-5.5-based diagnosis agent to identify physical failures in generated videos and assign each failure to physics categories, such as rigid-body motion, collision and fluid dynamics. Aggregating these results yields a category-level deficiency profile that captures the model's dominant physical weaknesses. We then utilize this profile to retrieve targeted training data from a large video gallery, where each candidate video is tagged with physics-domain labels derived from its caption, enabling retrieval that matches the deficient categories. 
The retrieved videos are added to the training set and re-captioned with our physics-aware pipeline in Sec.~\ref{sec:prompt_description}. In this way, language connects model diagnosis, data indexing, and targeted data acquisition, allowing the training distribution to be adapted to the generator's observed physical weaknesses.

\subsection{PhysThinker for Physics Reasoning} 
The VLM captioner and upsampler introduced in Sec.~\ref{sec:prompt_description} are initially instantiated with proprietary models. While these models provide strong physics reasoning capability, relying on them for large-scale video re-captioning and inference-time upsampling incurs high monetary cost.

To address this issue, we replace the costly GPT-based captioner and upsampler by distilling their physics reasoning capabilities into two efficient 4B vision-language models: \textbf{PhysThinker-C} for physics-aware video captioning and \textbf{PhysThinker-U} for inference-time prompt upsampling. PhysThinker-C learns from GPT-generated video-to-text annotations, while PhysThinker-U is trained to expand short text or image-text conditions into physics-rich prompts. Together, they provide scalable and low-cost alternatives to commercial models for large-scale annotation and inference. Training details of our PhysThinker are provided in App.~\ref{app:physthinker_train_detail}.

\section{Benchmark and Critic}
\label{sec:critic}

\subsection{Critics for Physics Caption}
\label{sec:critic—only}
\begin{figure}[t]
    \centering
    \includegraphics[width=\linewidth]{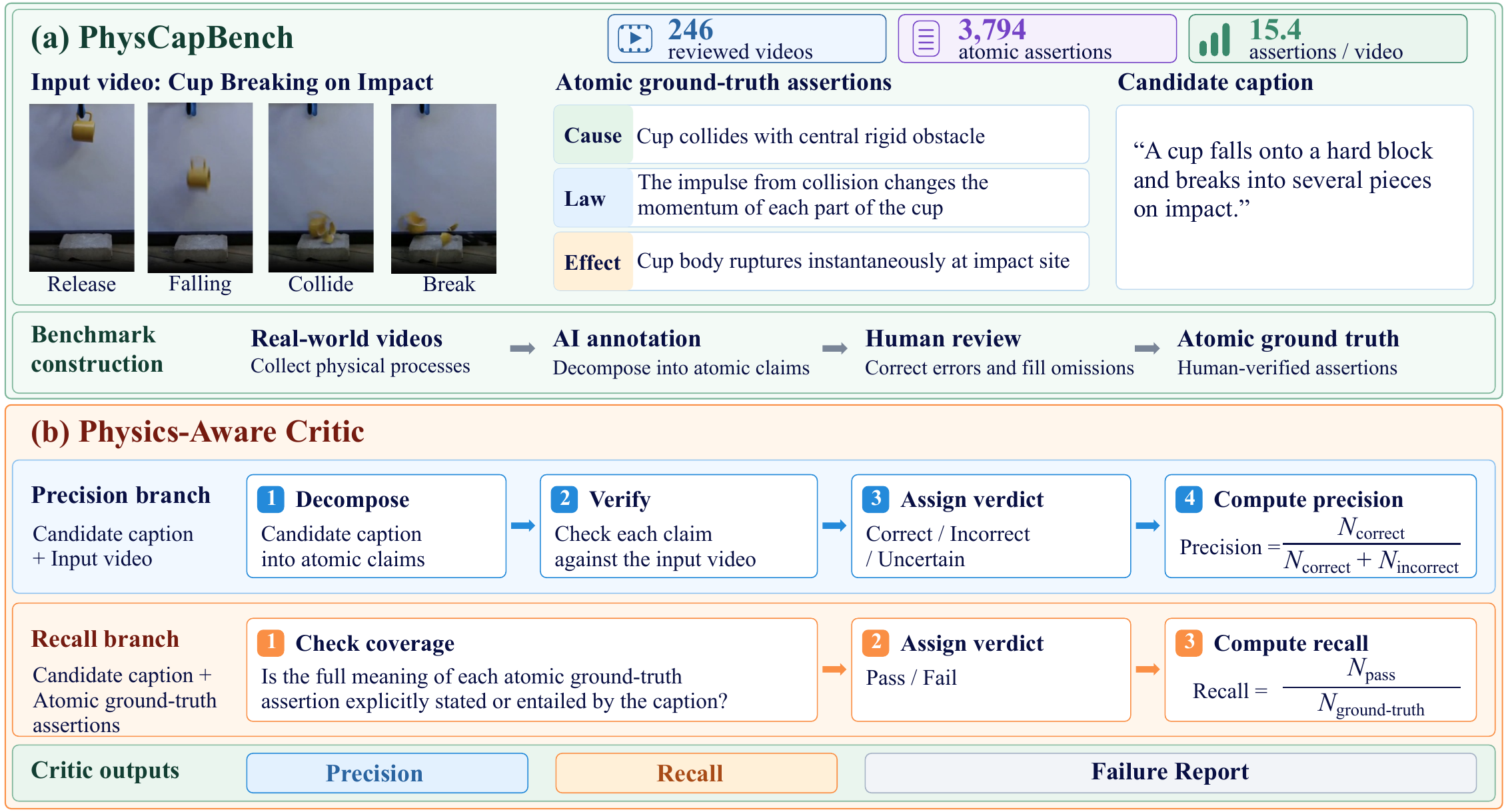}
    \vspace{-4mm}
    \caption{\textbf{Overview of our physics-aware critic and physics caption benchmark}. The critic evaluates generated captions from complementary precision and recall perspectives by verifying atomic claims against the input video and measuring the coverage of human-curated physical assertions. 
    }
    \vspace{-2mm}
    \label{fig:critic_overview}
\end{figure}

Fig.~\ref{fig:critic_overview}(b) illustrates the overview of our physics-aware critic used in the agentic loop. To systematically evaluate whether a video caption faithfully and comprehensively represents the physical dynamics in a video, inspired by the caption evaluation protocol used by Cosmos 3~\citep{nvidia2026cosmos3omnimodalworld}, we adopt \textbf{precision} and \textbf{recall} on the atomic assertions as critics, and specialize both metrics to physical content. Precision evaluates whether generated physical claims are visually supported by the video, while recall measures coverage of salient physical processes.

For evaluation of \textbf{precision}, the critic first decomposes the complete generated caption into atomic and independently verifiable claims. The critic is then given the full video together with each atomic claim and classifies it as \textit{correct}, \textit{incorrect}, or \textit{uncertain}. We compute micro-averaged precision as
$\mathrm{Precision}=N_{\mathrm{correct}}/(N_{\mathrm{correct}}+N_{\mathrm{incorrect}})$.
\textbf{Recall} instead focuses specifically on physical dynamics. The critic receives the generated caption together with a set of human-curated atomic physical assertions and determines whether each ground-truth assertion is sufficiently covered by the caption. An assertion receives a positive match only when the caption explicitly states or unambiguously entails its complete physical meaning. Recall is therefore computed as
$\mathrm{Recall}=N_{\mathrm{pass}}/N_{\mathrm{ground\text{-}truth}}$. We finally report the F1 score, defined as the harmonic mean of precision and recall, $\mathrm{F1}=2\cdot\mathrm{Precision}\cdot\mathrm{Recall}/(\mathrm{Precision}+\mathrm{Recall})$, as an overall measure of caption quality.

\subsection{Physics Caption Benchmark} 

\label{sec:physcapbench}

Based on the critic, we further introduce \textbf{PhysCapBench} (Fig.~\ref{fig:critic_overview}(a)), a dedicated benchmark for evaluating whether video captioning models can faithfully and comprehensively describe salient physical phenomena and their underlying dynamics, consisting of 246 physics-rich videos collected from Physics-IQ \citep{physicsiq2025} and YouTube.
We construct fine-grained physical annotations for each video using an agent-assisted, human-verified pipeline. AI annotators 
identify salient physical processes and decompose them into atomic assertions describing causes, relevant physical laws and effects. All candidate annotations are subsequently reviewed by humans, who remove incorrect or ambiguous assertions and supplement missing ones when necessary.

The final PhysCapBench contains \textbf{3,794 human-verified physical assertions} over 246 videos, with an average of 15.4 assertions per video. These annotations provide fine-grained supervision for evaluating whether a caption captures the key physical dynamics and reasoning in each video. Detailed implementations are provided in App.~\ref{app:physics-caption-benchmark}, and benchmark samples are listed in App.~\ref{sec:examples_physcapbench}.

\section{Experiment}
\label{sec:experiment}
\subsection{Experimental Details}
\label{sec:experimental_details}
\vspace{2pt}
\noindent \textbf{Dataset and Backbones.}
At the core of Physis-Lang is the construction of a physics-enriched video--language training dataset comprising 183K real-world videos. 
It combines 71K high-quality samples filtered from WISA-80K~\citep{wang2025wisa} with 112K additional videos selected through our deficiency-guided retrieval pipeline (Sec.~\ref{sec:video_retrieval}). 
We use the dataset to fine-tune several existing video-generation backbones without modifying their architectures and training objectives. Specifically, we consider Wan2.1-14B~\citep{wan2025}, for which the T2V-14B and I2V-14B variants are fine-tuned separately, and the Cosmos3 family~\citep{nvidia2026cosmos3omnimodalworld}, including Edge (4B), Nano (16B), and Super (64B), whose unified backbone supports both T2V and I2V generations. Further implementation details are provided in App.~\ref{sec:appendix_implementation_details}.

\vspace{2pt} \noindent \textbf{Baselines and Evaluation.} 
Our comparison involves cutting-edge pretrained models, including CogVideoX1.5-5B~\citep{yang2024cogvideox}, Wan2.2-TI2V-5B~\citep{wan2025}, HunyuanVideo-1.5~\citep{hunyuanvideo2025}, and the closed-source Veo~3.1. We also compare with other methods for promoting physical realism, including PhysVid~\citep{physvid2026}, PhyGDPO~\citep{phygdpo}, Self-Refinement~\citep{liu2025bootstrappingphysicsgroundedvideogeneration}, Kandinsky-WM~\citep{kandinskywm2026}, and PhiZero~\citep{shang2026phizero}. The evaluation is conducted on four widely used physical benchmarks: VideoPhy-2~\citep{videophy2_2025} and PhyGenBench~\citep{phygenbench2024} for Text-to-Video (T2V) generation, and PhyGround~\citep{lin2026phygroundbenchmarkingphysicalreasoning} and Physics-IQ Verified~\citep{rädsch2026physicsiqverified} for Image-to-Video (I2V) generation. 
To better distinguish differences in generation quality,
we replace the original offline VLM evaluators of VideoPhy-2
and PhyGenBench with GPT-5.5, which provides more
discriminative assessments in our comparisons
(App.~\ref{sec:gpt55}).
We evaluate all models using the same scoring protocol
within each benchmark to ensure fair comparisons.

\subsection{Comparison with State-of-the-Art Models}
\label{sec:sota_comparison}
Tables~\ref{tab:model_comparison_physicsiq}--\ref{tab:model_comparison_videophy} compare Cosmos3-Nano fine-tuned using our Physis-Lang with cutting-edge generative models and other baselines designed to improve physical realism in video generation.
For better comparison, we rescale each benchmark to a 100-point scoring scale.
We highlight three key advantages of our method: (i) \textbf{Consistent and significant improvement over base model.} Building on pretrained Cosmos3, Physis-Lang further improves its performance across all four evaluated physical video benchmarks. (ii) \textbf{State-of-the-art among open-source models.} Our method achieves the strongest performance among all open-source models and methods across the four benchmarks. (iii) \textbf{Outperforming the best closed-source model.} Our method surpasses Veo~3.1 on Physics-IQ Verified, PhyGround, and PhyGenBench, and remains competitive on VideoPhy-2, trailing by only 0.85 points on the full set (68.02 vs.\ 68.87) while outperforming it by 3.93 points on the hard split (62.36 vs.\ 58.43).
Qualitative examples in App.~\ref{sec:appendix_qualitative_comparison} demonstrate the improved generation quality of our method across the evaluated benchmarks, while App.~\ref{sec:demo_driving_robotics} further shows our strong performance in driving and robotics scenarios.
\begin{table*}[t]
\centering
\setlength{\tabcolsep}{2pt}
\renewcommand{\arraystretch}{1.0}

\begin{minipage}[t]{0.49\textwidth}
\centering
\scriptsize
\begin{minipage}[t][1.8\baselineskip][t]{\linewidth}
\captionof{table}{Physical fidelity on Physics-IQ Verified.}
\label{tab:model_comparison_physicsiq}
\end{minipage}\par
{\fontsize{7}{8.4}\selectfont

\begin{tabular*}{\linewidth}{@{\extracolsep{\fill}}lcccc>{\columncolor{tablegray}}c}
\toprule
\textbf{Model} & \textbf{ST} & \textbf{S} & \textbf{WS} & \textbf{MSE} & \cellcolor{white}\textbf{Final} \\
\midrule
CogVideoX1.5-5B~\tablecite{yang2024cogvideox} & 17.41 & 34.49 & 21.71 & 16.18 & 22.45 \\
Wan2.2-TI2V-5B~\tablecite{wan2025} & 23.26 & 33.89 & 23.18 & 23.19 & 25.88 \\
HunyuanVideo-1.5~\tablecite{hunyuanvideo2025} & 22.80 & 47.10 & 30.46 & 26.05 & 31.60 \\
Veo 3.1 & 19.01 & 52.16 & 35.45 & 33.32 & 34.99 \\
\midrule
Kandinsky-WM~\tablecite{kandinskywm2026} & 25.69 & 36.37 & 17.22 & 12.69 & 22.99 \\
Self-Refinement~\tablecite{liu2025bootstrappingphysicsgroundedvideogeneration} & 22.49 & 37.93 & 23.25 & 25.13 & 27.20 \\
PhiZero~\tablecite{shang2026phizero}  & 56.79 & 37.01 & 27.59 & 42.29 & \underline{40.91} \\
\midrule
Cosmos3-Nano & 29.38 & 51.76 & 38.92 & 40.87 & 40.23 \\
Ours (Cosmos3-Nano) & 33.59 & 55.38 & 42.45 & 42.23 & \textbf{43.41} \\
\bottomrule
\end{tabular*}%
}
\end{minipage}\hfill
\begin{minipage}[t]{0.49\textwidth}
\centering
\scriptsize
\begin{minipage}[t][1.8\baselineskip][t]{\linewidth}
\captionof{table}{Physical adherence on PhyGround.}
\label{tab:model_comparison_phyground}
\end{minipage}\par
{\fontsize{7}{8.4}\selectfont

\begin{tabular*}{\linewidth}{@{\extracolsep{\fill}}lcc>{\columncolor{tablegray}}c}
\toprule
\textbf{Model} & \textbf{General} & \textbf{Physics} & \cellcolor{white}\textbf{Overall} \\
\midrule
CogVideoX1.5-5B~\tablecite{yang2024cogvideox} & 59.02 & 58.36 & 58.68 \\
Wan2.2-TI2V-5B~\tablecite{wan2025} & 62.94 & 62.46 & 62.70 \\
HunyuanVideo-1.5~\tablecite{hunyuanvideo2025} & 67.14 & 65.42 & 66.28 \\
Veo 3.1 & 71.92 & 66.54 & \underline{69.24} \\
\midrule
Kandinsky-WM~\tablecite{kandinskywm2026} & 58.96 & 57.28 & 58.12 \\
Self-Refinement~\tablecite{liu2025bootstrappingphysicsgroundedvideogeneration} & 58.54 & 57.88 & 58.22\\
PhiZero~\tablecite{shang2026phizero}  & 52.27 & 63.42 & 57.85 \\
\midrule
Cosmos3-Nano & 65.46 & 64.90 & 65.18 \\
Ours (Cosmos3-Nano) & 71.54 & 68.26 & \textbf{69.90} \\
\bottomrule
\end{tabular*}%
}
\end{minipage}
\vspace{4pt}

\begin{minipage}[t]{0.49\textwidth}
\centering
\scriptsize
\begin{minipage}[t][1.8\baselineskip][t]{\linewidth}
\captionof{table}{Physical correctness on PhyGenBench.}
\label{tab:model_comparison_phygen}
\end{minipage}\par
{\fontsize{7}{8.4}\selectfont
\begin{tabular*}{\linewidth}{@{\extracolsep{\fill}}lcccc>{\columncolor{tablegray}}c}
\toprule
\textbf{Model} & \textbf{Force} & \textbf{Light} & \textbf{Heat} & \textbf{Material} & \cellcolor{white}\textbf{Overall} \\
\midrule
CogVideoX1.5-5B~\tablecite{yang2024cogvideox} & 38.33 & 57.33 & 55.56 & 43.33 & 48.75 \\
Wan2.2-TI2V-5B~\tablecite{wan2025} & 40.83 & 60.67 & 43.33 & 40.83 & 47.50 \\
HunyuanVideo-1.5~\tablecite{hunyuanvideo2025} & 47.50 & 56.67 & 44.44 & 41.67 & 48.33 \\
Veo 3.1 & 65.00 & 70.67 & 57.78 & 65.83 & \underline{65.63} \\
\midrule
PhysVid~\tablecite{physvid2026} & 48.33 & 62.00 & 33.33 & 40.83 & 47.92 \\
PhyGDPO~\tablecite{phygdpo} & 43.33 & 62.67 & 43.33 & 41.67 & 48.96 \\
Self-Refinement~\tablecite{liu2025bootstrappingphysicsgroundedvideogeneration} & 46.67 & 59.33 & 44.44 & 42.50 & 49.17 \\
\midrule
Cosmos3-Nano & 64.17 & 64.00 & 57.78 & 59.17 & 61.67 \\
Ours (Cosmos3-Nano) & 67.50 & 72.67 & 75.56 & 69.17 & \textbf{71.04} \\
\bottomrule
\end{tabular*}%
}
\end{minipage}\hfill
\begin{minipage}[t]{0.49\textwidth}
\centering
\scriptsize
\begin{minipage}[t][1.8\baselineskip][t]{\linewidth}
\captionof{table}{Physical commonsense on VideoPhy-2.}
\label{tab:model_comparison_videophy}
\end{minipage}\par
{\fontsize{7}{8.4}\selectfont
\begin{tabular*}{\linewidth}{@{\extracolsep{\fill}}lcccc>{\columncolor{tablegray}}c>{\columncolor{tablegray}}c}
\toprule
\textbf{Model} & \textbf{SA} & \textbf{PC} & \textbf{SA$\geq$4} & \textbf{PC$\geq$4} & \cellcolor{white}\textbf{All} & \cellcolor{white}\textbf{Hard} \\
\midrule
CogVideoX1.5-5B~\tablecite{yang2024cogvideox} & 3.74 & 4.10 & 63.79 & 74.79 & 49.41 & 33.15 \\
Wan2.2-TI2V-5B~\tablecite{wan2025} & 3.69 & 4.39 & 60.91 & 85.45 & 53.98 & 42.13 \\
HunyuanVideo-1.5~\tablecite{hunyuanvideo2025} & 3.84 & 4.57 & 67.85 & 92.39 & 62.94 & 51.69 \\
Veo 3.1 & 4.39 & 4.06 & 84.43 & 80.54 & \textbf{68.87} & \underline{58.43} \\
\midrule
PhysVid~\tablecite{physvid2026} & 3.50 & 4.27 & 55.33 & 82.74 & 48.22 & 29.78 \\
PhyGDPO~\tablecite{phygdpo} & 3.74 & 4.64 & 62.27 & 93.57 & 59.56 & 44.94 \\
Self-Refinement~\tablecite{liu2025bootstrappingphysicsgroundedvideogeneration} & 3.60 & 4.26 & 56.01 & 82.91 & 47.88 & 28.09 \\
\midrule
Cosmos3-Nano & 3.86 & 4.49 & 66.33 & 90.19 & 60.41 & 48.31 \\
Ours (Cosmos3-Nano) & 4.09 & 4.44 & 75.97 & 89.00 & \underline{68.02} & \textbf{62.36} \\
\bottomrule
\end{tabular*}%
}
\end{minipage}
\end{table*}

\subsection{Generalization across Backbones and Scales}
\label{sec:backbone_scale}
Tab.~\ref{tab:backbone_scale} evaluates whether our method
generalizes across model families and parameter scales.
We draw three conclusions:
(i) \textbf{Consistent gains across model families.}
Our method improves Wan2.1-14B and Cosmos3-Nano-16B by
7.05 and 6.22 points on average, respectively, with gains
on all four physical benchmarks.
(ii) \textbf{Effectiveness across model scales.}
Within the Cosmos~3 family, our method yields average gains
of 3.24, 6.22, and 5.02 points on Edge-4B, Nano-16B, and
Super-64B, respectively. These results demonstrate benefits
across a broad range of model sizes.
(iii) \textbf{Preserved general video quality.}
Our fine-tuning strategy does not compromise general video generation quality; please refer to App.~\ref{sec:vbench} for more details.
\begin{table}[t]
\centering\footnotesize
\renewcommand{\arraystretch}{0.88}
\caption{\textbf{Effectiveness on different backbones.} Each cell shows the improvement brought by our method. Mean $\Delta$ is the average improvement across the four benchmarks. }
\label{tab:backbone_scale}
\setlength{\tabcolsep}{4pt}
\resizebox{0.94\linewidth}{!}{\begin{tabular}{lccccc}
\toprule
\textbf{Generator} & \textbf{PhyGenBench} & \textbf{Physics-IQ Verified} & \textbf{VideoPhy-2} & \textbf{PhyGround} & \textbf{Mean $\Delta$}\\
\midrule
\multicolumn{6}{l}{\textit{Across model families}}\\
Wan2.1-14B & 56.67 $\rightarrow$ 65.83 & 27.87 $\rightarrow$ 35.15 & 57.02 $\rightarrow$ 65.65 & 61.52 $\rightarrow$ 64.64 & +7.05\\
Cosmos3-Nano-16B & 61.67 $\rightarrow$ 71.04 & 40.23 $\rightarrow$ 43.41 & 60.41 $\rightarrow$ 68.02 & 65.18 $\rightarrow$ 69.90 & +6.22\\
\midrule
\multicolumn{6}{l}{\textit{Across Cosmos scales}}\\
Cosmos3-Edge-4B & 48.96 $\rightarrow$ 52.50 & 32.80 $\rightarrow$ 34.69 & 32.99 $\rightarrow$ 40.61 & 66.66 $\rightarrow$ 66.56 & +3.24\\
Cosmos3-Nano-16B & 61.67 $\rightarrow$ 71.04 & 40.23 $\rightarrow$ 43.41 & 60.41 $\rightarrow$ 68.02 & 65.18 $\rightarrow$ 69.90 & +6.22\\
Cosmos3-Super-64B & 66.04 $\rightarrow$ 70.21 & 45.92 $\rightarrow$ 50.00 & 60.91 $\rightarrow$ 72.42 & 68.67 $\rightarrow$ 68.98 & +5.02\\
\bottomrule
\end{tabular}}
\end{table}

\subsection{Ablation and Analysis}
\label{sec:ablation_analysis}
\subsubsection{Agentic Caption Evolution}
\label{sec:loop_effectiveness}
\begin{figure}[t]
\centering
\includegraphics[width=0.86\linewidth]{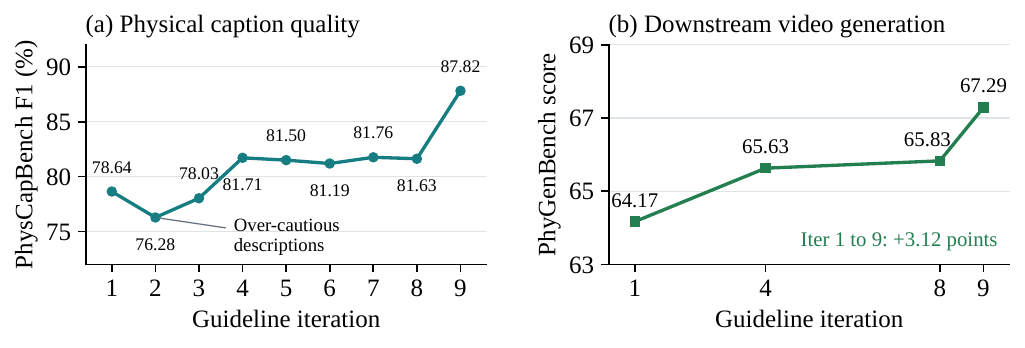}
\vspace{-2mm}
\caption{\textbf{From caption evolution to video generation.} Left: PhysCapBench F1 at different iterations. Right: physics-aware video generation performance of the same pretrained Cosmos3 model with inference captions produced at different iterations. }
\label{fig:loop_transfer}
\end{figure}

Fig.~\ref{fig:loop_transfer} evaluates whether caption
evolution improves both physical descriptions and downstream
video generation. We highlight two observations:
(i) \textbf{The agentic loop improves physical caption quality.}
On PhysCapBench, the F1 score increases from 78.64 at Iteration 1 to
87.82 at Iteration 9, showing the effectiveness of critic-guided guideline refinement.
(ii) \textbf{The self-evolving guidelines improve video generation.}
Keeping the pretrained Cosmos3-Nano fixed, we change
only the inference captions produced using guidelines from
Iterations 1, 4, 8 and 9. PhyGenBench scores increase from
64.17 to 65.63, 65.83 and 67.29, respectively.
App.~\ref{sec:loop_full} provides the complete evolution
trajectory. App.~\ref{sec:analysis_attention} analyses the effect of \textit{physics\_reasoning} derived from our agentic loop.

\subsubsection{Language-Guided Data Curation}
\label{sec:data_curation_eval}

\begin{figure*}[t]
\centering
\begin{minipage}[t]{0.42\textwidth}
\vspace{0pt}
\centering\footnotesize
\captionof{table}{\textbf{Effect of retrieved training data.} Benchmark scores before and after data expansion.}
\label{tab:retrieval_effect}
\setlength{\tabcolsep}{2.5pt}
\resizebox{\linewidth}{!}{%
\begin{tabular}{lrrr}
\toprule
\textbf{Benchmark} & \textbf{WISA} & \textbf{+ Retrieved} & \textbf{Gain}\\
\midrule
PhyGenBench & 68.12 & 71.04 & +2.92\\
Physics-IQ Verified & 40.68 & 43.41 & +2.73\\
VideoPhy-2 & 64.63 & 68.02 & +3.39\\
\midrule
Mean & 57.81 & 60.82 & +3.01\\
\bottomrule
\end{tabular}}
\end{minipage}\hfill
\begin{minipage}[t]{0.55\textwidth}
\vspace{-3pt}
\centering
\includegraphics[width=\linewidth,trim=0 3bp 0 4bp,clip]{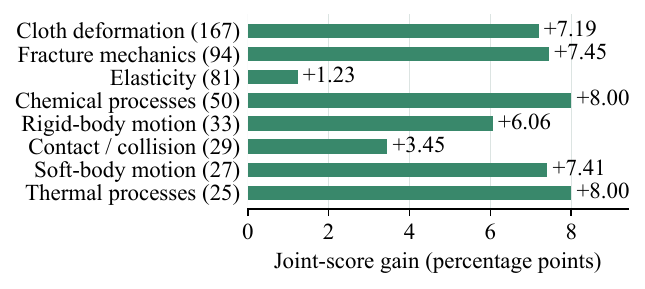}\par
\captionsetup{position=bottom,skip=5pt}
\caption{\textbf{VideoPhy-2 category gains} after retrieval. }
\label{fig:retrieval_category_gains}
\end{minipage}
\end{figure*}

\textbf{Language-guided data curation (Sec.~\ref{sec:video_retrieval}) consistently improves physical generation.} Adding retrieved videos to the training set improves benchmarks, for an average gain of 3.01 points (Tab.~\ref{tab:retrieval_effect}). To further examine whether these gains are consistent across our physical tags, 
we break down the improvements on VideoPhy-2 by physical category (a sample may belong to multiple categories). The gains are positive across all frequently represented categories, such as chemical processes (+8.00), fracture mechanics (+7.45), and cloth deformation (+7.19 points) (Fig.~\ref{fig:retrieval_category_gains}).

\subsubsection{Prompting, Fine-Tuning, and Negative Guidance}
\label{sec:prompt_components}

Tab.~\ref{tab:prompt_components} highlights three findings:
(i) \textbf{Physical prompting is effective without fine-tuning.} Our physical reasoning prompt improves the pretrained model from 61.67 to 63.33 ($\mathbb{A}\rightarrow\mathbb{C}$), demonstrating a zero-shot gain. Moreover, our agentic loop further improves over the manually designed prompt~\citep{phygdpo}($\mathbb{B}\rightarrow\mathbb{C}$), validating the effectiveness. (ii) \textbf{SFT provides additional gains.} With inference prompts fixed, fine-tuning brings further improvement: 3.55 points ($\mathbb{C}\rightarrow\mathbb{G}$) and 3.75 points ($\mathbb{D}\rightarrow\mathbb{H}$). (iii) \textbf{Negative guidance complements SFT.} Adding physics negative prompts improves the fine-tuned model by 2.50 ($\mathbb{E}\rightarrow\mathbb{F}$) and 4.16 points ($\mathbb{G}\rightarrow\mathbb{H}$), respectively.

\begin{table}[t]
\centering
\footnotesize
\setlength{\tabcolsep}{5pt}
\caption{\textbf{Prompting and SFT on PhyGenBench.}
Base: base caption;
M: manually designed physics caption;
P: \textbf{\textit{physics\_reasoning}};
N: \textbf{\textit{physics\_negative\_prompt}}.}
\label{tab:prompt_components}
\begin{tabular}{lllr@{\hspace{14pt}}lllr}
\toprule
\textbf{ID} & \textbf{SFT data} & \textbf{Prompt}
& \textbf{Score}
& \textbf{ID} & \textbf{SFT data} & \textbf{Prompt}
& \textbf{Score}\\
\midrule
$\mathbb{A}$ & None & Base & 61.67
& $\mathbb{E}$ & WISA & Base + P & 65.62\\
$\mathbb{B}$ & None & M & 61.86
& $\mathbb{F}$ & WISA & Base + P + N & 68.12\\
$\mathbb{C}$ & None & Base + P & 63.33
& $\mathbb{G}$ & WISA + retrieved & Base + P & 66.88\\
$\mathbb{D}$ & None & Base + P + N & \textbf{67.29}
& $\mathbb{H}$ & WISA + retrieved
& Base + P + N & \textbf{71.04}\\
\bottomrule
\end{tabular}
\end{table}

\subsection{Open-Source Deployment with PhysThinker}
\label{sec:physthinker_eval}
We train PhysThinker-C and PhysThinker-U for video captioning and inference-time prompt upsampling, respectively. Tab.~\ref{tab:physthinker_replacement} demonstrates an accuracy--cost trade-off: commercial, captioner-only replacement, and fully local pipelines improve over the pretrained baseline by 7.05, 6.76, and 4.76 points on average, while estimated external annotation costs decrease from \$24.12K to \$0.12K and \$0.
Costs cover the re-captioning of training data and the upsampling for benchmarks, whereas accuracy is averaged on four reported benchmarks.
\begin{table}[!htbp]
\centering\small
\caption{\textbf{PhysThinker replacement on Wan2.1-14B.} VideoPhy-2 reports All / Hard; 
Cost denotes API expenditure for
training-data captioning and benchmark upsampling (USD).
}
\vspace{-2mm}
\label{tab:physthinker_replacement}
\setlength{\tabcolsep}{3pt}
\resizebox{\linewidth}{!}{\begin{tabular}{llrrrrrr}
\toprule
\textbf{Captioner} & \textbf{Upsampler} & \textbf{PhyGenBench} & \textbf{Physics-IQ Verified} & \textbf{VideoPhy-2} & \textbf{PhyGround} & \textbf{Mean $\Delta$} & \textbf{Cost (USD)}\\
\midrule
\multicolumn{2}{l}{Pretrained Wan2.1-14B} & 56.67 & 27.87 & 57.02 / 43.82 & 61.52 & -- & --\\
\midrule
Commercial & Commercial & 65.83 & 35.15 & 65.65 / 59.55 & 64.64 & +7.05 & $\sim$24.12K\\
PhysThinker-C & Commercial & 65.00 & 34.71 & 65.99 / 54.49 & 64.40 & +6.76 & $\sim$0.12K\\
PhysThinker-C & PhysThinker-U & 63.33 & 34.13 & 60.07 / 55.06 & 64.60 & +4.76 & 0\\
\bottomrule
\end{tabular}}
\vspace{-2mm}
\end{table}

\section{Conclusion}
\label{sec:conclusion}

Video world models require physical plausibility beyond visual quality. We introduce \textit{Physis-Lang}, which treats structured physical language as a shared representation across data curation, model training, and inference. PhysCapBench and an agentic self-evolution loop refine physical captions, while language-guided retrieval targets missing physical processes. Experiments with Wan2.1 and Cosmos3 family models
across four physical benchmarks of video generation show consistent and significant improvements over the pretrained models. 
These results show that language, when explicitly structured and self-evolving, can serve not merely as a conditioning interface, but as a practical representation for carrying, refining, and applying physical knowledge in video world models.

{
  \small
  \bibliographystyle{unsrt}
  \bibliography{arxiv}

@article{physhpo2025,
  title={Hierarchical Fine-grained Preference Optimization for Physically Plausible Video Generation},
  author={Chen, Harold Haodong and Huang, Haojian and Chen, Qifeng and Yang, Harry and Lim, Ser-Nam},
  journal={Advances in Neural Information Processing Systems},
  volume={38},
  pages={148434--148466},
  year={2025}
}

@inproceedings{pisa2025,
  title={{PISA} Experiments: Exploring Physics Post-Training for Video Diffusion Models by Watching Stuff Drop},
  author={Li, Chenyu and Michel, Oscar and Pan, Xichen and Liu, Sainan and Roberts, Mike and Xie, Saining},
  booktitle={International Conference on Machine Learning},
  pages={35685--35709},
  year={2025},
  organization={PMLR}
}

@inproceedings{tgt2025,
  title={{TGT}: Text-Grounded Trajectories for Locally Controlled Video Generation},
  author={Zhang, Guofeng and Wang, Angtian and Fang, Jacob Zhiyuan and Jiang, Liming and Yang, Haotian and Liu, Bo and Yang, Yiding and Chen, Guang and Wen, Longyin and Yuille, Alan and Ma, Chongyang},
  booktitle={Proceedings of the IEEE/CVF Conference on Computer Vision and Pattern Recognition},
  pages={22028--22037},
  year={2026}
}

@inproceedings{chainofevent2026,
  title={Chain of Event-Centric Causal Thought for Physically Plausible Video Generation},
  author={Wang, Zixuan and Hu, Yixin and Wang, Haolan and Chen, Feng and Liu, Yan and Li, Wen and Lei, Yinjie},
  booktitle={Proceedings of the IEEE/CVF Conference on Computer Vision and Pattern Recognition},
  pages={38122--38131},
  year={2026}
}

@inproceedings{phantom2026,
  title={{PHANTOM}: Physics-Infused Video Generation via Joint Modeling of Visual and Latent Physical Dynamics},
  author={Shen, Ying and Xiong, Jerry and Yu, Tianjiao and Lourentzou, Ismini},
  booktitle={Proceedings of the IEEE/CVF Conference on Computer Vision and Pattern Recognition},
  pages={11185--11194},
  year={2026}
}

@inproceedings{moregen2026,
  title={{MoReGen}: Multi-Agent Motion-Reasoning Engine for Code-based Text-to-Video Synthesis},
  author={Bai, Xiangyu and Liang, He and Galoaa, Bishoy and Nandi, Utsav and Moezzi, Shayda and He, Yuhang and Ostadabbas, Sarah},
  booktitle={Proceedings of the IEEE/CVF Conference on Computer Vision and Pattern Recognition},
  pages={7632--7642},
  year={2026}
}

@article{reasoningimplausibility2025,
  title={Enhancing Physical Plausibility in Video Generation by Reasoning the Implausibility},
  author={Hao, Yutong and Chen, Chen and Mian, Ajmal Saeed and Xu, Chang and Liu, Daochang},
  journal={arXiv preprint arXiv:2509.24702},
  year={2025}
}

@inproceedings{physinone2026,
  title={{PhysInOne}: Visual Physics Learning and Reasoning in One Suite},
  author={Zhou, Siyuan and Wang, Hejun and Cheng, Hu and Li, Jinxi and Wang, Dongsheng and Jiang, Junwei and Jin, Yixiao and Huang, Jiayue and Mao, Shiwei and Liu, Shangjia and Yang, Yafei and Song, Hongkang and Wei, Shenxing and Zhang, Zihui and Wang, Bing and Wang, Zhihua and Zou, Chuhang and Yang, Bo},
  booktitle={Proceedings of the IEEE/CVF Conference on Computer Vision and Pattern Recognition},
  pages={33131--33142},
  year={2026}
}

@inproceedings{tsa2026,
  title={Tempered Self-Similarity Alignment for Physically Plausible Video Generation},
  author={Kim, Manjin and Kwak, Suha and Cho, Minsu},
  booktitle={Proceedings of the IEEE/CVF Conference on Computer Vision and Pattern Recognition Workshops},
  pages={5148--5158},
  year={2026}
}

@article{newton2026,
  title={NEWTON: Agentic Planning for Physically Grounded Video Generation},
  author={Feng, Yuxiang and Wang, Juncheng and Xu, Chao and Qian, Yijie and Wang, Huihan and Hou, Wenlong and Liu, Yang and Sun, Baigui and Liu, Yong and Wang, Shujun},
  journal={arXiv preprint arXiv:2605.18396},
  year={2026}
}

@inproceedings{dynamicsboost2026,
  title={{DynamicsBoost}: Dynamic Plausible Video Generation via Annotation-Free Continuation Preference Optimization},
  author={Li, Jiaxing and Wang, Jiepeng and Gao, Junyao and Liu, Yang and Li, Eric and An, Bo and Guo, Hao-Xiang},
  booktitle={Proceedings of the IEEE/CVF Conference on Computer Vision and Pattern Recognition},
  pages={20024--20033},
  year={2026}
}

@article{ditmem2025,
  title={Learning Plug-and-play Memory for Guiding Video Diffusion Models},
  author={Song, Selena and Xu, Ziming and Zhang, Zijun and Zhou, Kun and Guo, Jiaxian and Qin, Lianhui and Huang, Biwei},
  journal={arXiv preprint arXiv:2511.19229},
  year={2025}
}

@article{direct2026,
  title={DiReCT: Disentangled Regularization of Contrastive Trajectories for Physics-Refined Video Generation},
  author={Meyarian, Abolfazl and Monsefi, Amin Karimi and Ramnath, Rajiv and Lim, Ser-Nam},
  journal={arXiv preprint arXiv:2603.25931},
  year={2026}
}

@article{phyprompt2026,
  title={PhyPrompt: RL-based Prompt Refinement for Physically Plausible Text-to-Video Generation},
  author={Wu, Shang and Xu, Chenwei and Xia, Zhuofan and Li, Weijian and Lu, Lie and Maneriker, Pranav and Du, Fan and Li, Manling and Liu, Han},
  journal={arXiv preprint arXiv:2603.03505},
  year={2026}
}

@article{motionforcing2026,
  title={Motion Forcing: A Decoupled Framework for Robust Video Generation in Motion Dynamics},
  author={Xu, Tianshuo and Chen, Zhifei and Wu, Leyi and Lu, Hao and Chen, Ying-cong},
  journal={arXiv preprint arXiv:2603.10408},
  year={2026}
}

@article{mmphysvideo2026,
  title={MMPhysVideo: Physically Plausible Video Generation Through Joint RGB-Perception Modeling},
  author={Lin, Shubo and Zhang, Xuanyang and Cheng, Wei and Hu, Weiming and Yu, Gang and Gao, Jin},
  journal={arXiv preprint arXiv:2604.02817},
  year={2026}
}

@article{causalmotion2026,
  title={CausalMotion: Structured Physical Reasoning as Keyframe and Trajectory Guidance for Training-Free Video Generation},
  author={Zhuang, Sihan and Chen, Xinyuan and Xue, Tianfan and Wang, Yaohui},
  journal={arXiv preprint arXiv:2606.14317},
  year={2026}
}

@article{diffusiondrf2026,
  title={Diffusion-DRF: Free, Rich, and Differentiable Reward for Video Diffusion Fine-Tuning},
  author={Wang, Yifan and Li, Yanyu and Qian, Gordon Guocheng and Tulyakov, Sergey and Fu, Yun and Kag, Anil},
  journal={arXiv preprint arXiv:2601.04153},
  year={2026}
}

@inproceedings{newtongen2026,
  title={{NewtonGen}: Physics-Consistent and Controllable Text-to-Video Generation via Neural {N}ewtonian Dynamics},
  author={Yuan, Yu and Wang, Xijun and Wickremasinghe, Tharindu and Nadir, Zeeshan and Ma, Bole and Chan, Stanley H.},
  booktitle={International Conference on Learning Representations},
  year={2026},
  url={https://openreview.net/forum?id=rJ6N6sunaU}
}

@article{seekingphysics2026,
  title={Seeking Physics in Diffusion Noise},
  author={Tang, Chujun and Zhong, Lei and Ding, Fangqiang},
  journal={arXiv preprint arXiv:2603.14294},
  year={2026}
}

@article{physcorr2025,
  title={PhysCorr: Dual-Reward DPO for Physics-Constrained Text-to-Video Generation with Automated Preference Selection},
  author={Wang, Peiyao and Wang, Weining and Li, Qi},
  journal={arXiv preprint arXiv:2511.03997},
  year={2025}
}

@article{phyworld2026,
  title={PhyWorld: Physics-Faithful World Model for Video Generation},
  author={Zhao, Pu and Lin, Juyi and Rupprecht, Timothy and Akbari, Arash and Yang, Chence and Chowdhury, Rahul and Motamedi, Elaheh and Akbari, Arman and He, Yumei and Wang, Chen and Yuan, Geng and Chen, Weiwei and Wang, Yanzhi},
  journal={arXiv preprint arXiv:2605.19242},
  year={2026}
}

@inproceedings{wmreward2026,
  title={Inference-time Physics Alignment of Video Generative Models with Latent World Models},
  author={Yuan, Jianhao and Zhang, Xiaofeng and Friedrich, Felix and Beltran-Velez, Nicolas and Hall, Melissa and Askari-Hemmat, Reyhane and Han, Xiaochuang and Ballas, Nicolas and Drozdzal, Michal and Romero-Soriano, Adriana},
  booktitle={Proceedings of the IEEE/CVF Conference on Computer Vision and Pattern Recognition},
  pages={16118--16129},
  year={2026}
}

@article{physvideogenerator2026,
  title={PhysVideoGenerator: Towards Physically Aware Video Generation via Latent Physics Guidance},
  author={Satish, Siddarth Nilol Kundur and Jaiswal, Devesh and Chen, Hongyu and Bakshi, Abhishek},
  journal={arXiv preprint arXiv:2601.03665},
  year={2026}
}

@inproceedings{prophy2026,
  title={{ProPhy}: Progressive Physical Alignment for Dynamic World Simulation},
  author={Wang, Zijun and Hu, Panwen and Wang, Jing and Zhang, Terry Jingchen and Cheng, Yuhao and Chen, Long and Yan, Yiqiang and Jiang, Zutao and Li, Hanhui and Liang, Xiaodan},
  booktitle={Proceedings of the IEEE/CVF Conference on Computer Vision and Pattern Recognition},
  pages={14492--14501},
  year={2026}
}

@inproceedings{physvid2026,
  title={{PhysVid}: Physics Aware Local Conditioning for Generative Video Models},
  author={Pathak, Saurabh and Arani, Elahe and Pechenizkiy, Mykola and Zonooz, Bahram},
  booktitle={Proceedings of the IEEE/CVF Conference on Computer Vision and Pattern Recognition},
  pages={41847--41858},
  year={2026}
}

@inproceedings{moalign2025,
  title={{MoAlign}: Motion-Centric Representation Alignment for Video Diffusion Models},
  author={Bhowmik, Aritra and Korzhenkov, Denis and Snoek, Cees G. M. and Habibian, Amirhossein and Ghafoorian, Mohsen},
  booktitle={International Conference on Learning Representations},
  year={2026},
  url={https://openreview.net/forum?id=OR0ySm4l9h}
}

@article{stance2025,
  title={STANCE: Motion Coherent Video Generation Via Sparse-to-Dense Anchored Encoding},
  author={Chen, Zhifei and Xu, Tianshuo and Wu, Leyi and Wang, Luozhou and Yan, Dongyu and You, Zihan and Luo, Wenting and Zhang, Guo and Chen, Yingcong},
  journal={arXiv preprint arXiv:2510.14588},
  year={2025}
}

@article{phdreamer2026,
  title={PH-Dreamer: A Physics-Driven World Model via Port-Hamiltonian Generative Dynamics},
  author={Luan, Xueyu and Shi, Chenwei},
  journal={arXiv preprint arXiv:2605.18303},
  year={2026}
}

@inproceedings{envisioningfuture2026,
  title={Envisioning the Future, One Step at a Time},
  author={Baumann, Stefan Andreas and Wiese, Jannik and Martorella, Tommaso and Kalayeh, Mahdi M. and Ommer, Bj{\"o}rn},
  booktitle={Proceedings of the IEEE/CVF Conference on Computer Vision and Pattern Recognition},
  pages={6823--6836},
  year={2026}
}

@inproceedings{longtermmotion2026,
  title={Learning Long-term Motion Embeddings for Efficient Kinematics Generation},
  author={Stracke, Nick and Bauer, Kolja and Baumann, Stefan Andreas and Bautista, Miguel {\'A}ngel and Susskind, Josh and Ommer, Bj{\"o}rn},
  booktitle={Proceedings of the IEEE/CVF Conference on Computer Vision and Pattern Recognition},
  pages={42581--42591},
  year={2026}
}

@inproceedings{pris2026,
  title={Rethinking Prompt Design for Inference-time Scaling in Text-to-Visual Generation},
  author={Kim, Subin and Mo, Sangwoo and Rizve, Mamshad Nayeem and Xu, Yiran and Liu, Difan and Shin, Jinwoo and Hinz, Tobias},
  booktitle={Proceedings of the IEEE/CVF Conference on Computer Vision and Pattern Recognition},
  pages={22090--22099},
  year={2026}
}

@inproceedings{latsearch2026,
  title={{LatSearch}: Latent Reward-Guided Search for Faster Inference-Time Scaling in Video Diffusion},
  author={Zhao, Zengqun and Liu, Ziquan and Cao, Yu and Gong, Shaogang and Zhang, Zhensong and Song, Jifei and Deng, Jiankang and Patras, Ioannis},
  booktitle={European Conference on Computer Vision},
  pages={170--189},
  year={2026},
  organization={Springer}
}

@article{diffusionapo2026,
  title={Diffusion-APO: Trajectory-Aware Direct Preference Alignment for Video Diffusion Transformers},
  author={Zhu, Jingyuan and Chen, Biaolong and Zhang, Le and Zhang, Aixi and Jiang, Hao and Huang, Pipei},
  journal={arXiv preprint arXiv:2605.07503},
  year={2026}
}

@inproceedings{vchain2025,
  title={{VC}hain: Chain-of-Visual-Thought for Reasoning in Video Generation},
  author={Huang, Ziqi and Yu, Ning and Chen, Gordon and Qiu, Haonan and Debevec, Paul and Liu, Ziwei},
  booktitle={Findings of the Association for Computational Linguistics: ACL 2026},
  pages={226--250},
  year={2026}
}

@article{diffphy2025,
  title={Think Before You Diffuse: Infusing Physical Rules into Video Diffusion},
  author={Zhang, Ke and Xiao, Cihan and Xu, Jiacong and Mei, Yiqun and Patel, Vishal M.},
  journal={arXiv preprint arXiv:2505.21653},
  year={2025}
}

@inproceedings{ragme2025,
  title={{RagMe}: Retrieval Augmented Video Generation for Enhanced Motion Realism},
  author={Peruzzo, Elia and Xu, Dejia and Xu, Xingqian and Shi, Humphrey and Sebe, Nicu},
  booktitle={Proceedings of the 2025 International Conference on Multimedia Retrieval},
  pages={1081--1090},
  year={2025}
}

@article{motionrag2025,
  title={{MotionRAG}: Motion Retrieval-Augmented Image-to-Video Generation},
  author={Zhu, Chenhui and Wu, Yilu and Wang, Shuai and Wu, Gangshan and Wang, Limin},
  journal={Advances in Neural Information Processing Systems},
  volume={38},
  pages={192087--192110},
  year={2025}
}

@inproceedings{physrag2026,
  title={{PhysRAG}: Enhancing Physics-Awareness in Video Generation via Retrieval-Augmented Generation},
  author={Cheng, Kexu and Liu, Zicheng and Gao, Mingju and Song, Chunhe and Tang, Hao},
  booktitle={European Conference on Computer Vision},
  pages={558--578},
  year={2026},
  organization={Springer}
}

@article{phyparam2026,
  title={Learning Explicit Physical Parameter Control and Benchmarking for Video Generation},
  author={Li, Yanxun and Wen, Hao and Song, Bingze and Zhu, Jiashu and Hao, Aiming and Chen, Chubin and Chen, Jintao and Wu, Jiahong and Chu, Xiangxiang and Wang, Miao},
  journal={arXiv preprint arXiv:2607.18924},
  year={2026}
}

@article{psdpo2026,
  title={When Physical Preferences Meet Semantic Constraints: Physical and Semantic Direct Preference Optimization for Text-to-Video Generation},
  author={Meng, Siwei and Luo, Yawei and Zhang, Shu and Liu, Ping},
  journal={arXiv preprint arXiv:2607.16947},
  year={2026},
  note={Accepted to ACM Multimedia 2026}
}

@article{videococo2026,
  title={VideoCoCo: Code-as-CoT for Physically-Consistent Video Generation via an Agentic Dual-Engine System},
  author={Li, Haodong and Ren, Tianfei and Ma, Xiaoxiao and Qing, Chunmei and Fang, Zhen and He, Sipeng and Guo, Ziyu and Wu, Haoyu and Tian, Juanxi and Zou, Yihang and An, Ruichuan and Jiang, Dongzhi and Yang, Boxue and Xie, Ji and Huang, Xu and Yan, Wenhao and Zou, Jialv and Yue, Zhengrong and Luo, Yaxin and Li, Xiaotong and Wang, Yuzhu and Ye, Junyan and Zhao, Jinjing and Chen, Zehui and Chen, Lin and Yan, Renye and Zhao, Feng and Heng, Pheng-Ann},
  journal={arXiv preprint arXiv:2607.27380},
  year={2026}
}

@inproceedings{selfinstruct2023,
  title={Self-Instruct: Aligning Language Models with Self-Generated Instructions},
  author={Wang, Yizhong and Kordi, Yeganeh and Mishra, Swaroop and Liu, Alisa and Smith, Noah A. and Khashabi, Daniel and Hajishirzi, Hannaneh},
  booktitle={Proceedings of the 61st Annual Meeting of the Association for Computational Linguistics (Volume 1: Long Papers)},
  pages={13484--13508},
  year={2023}
}

@inproceedings{wizardlm2023,
  title={{WizardLM}: Empowering Large Pre-Trained Language Models to Follow Complex Instructions},
  author={Xu, Can and Sun, Qingfeng and Zheng, Kai and Geng, Xiubo and Zhao, Pu and Feng, Jiazhan and Tao, Chongyang and Lin, Qingwei and Jiang, Daxin},
  booktitle={International Conference on Learning Representations},
  year={2024},
  url={https://openreview.net/forum?id=CfXh93NDgH}
}

@inproceedings{selfrewarding2024,
  title={Self-Rewarding Language Models},
  author={Yuan, Weizhe and Pang, Richard Yuanzhe and Cho, Kyunghyun and Li, Xian and Sukhbaatar, Sainbayar and Xu, Jing and Weston, Jason E.},
  booktitle={International Conference on Machine Learning},
  pages={57905--57923},
  year={2024},
  organization={PMLR}
}

@article{ragen2025,
  title={RAGEN: Understanding Self-Evolution in LLM Agents via Multi-Turn Reinforcement Learning},
  author={Wang, Zihan and Wang, Kangrui and Wang, Qineng and Zhang, Pingyue and Li, Linjie and Yang, Zhengyuan and Jin, Xing and Yu, Kefan and Nguyen, Minh Nhat and Liu, Licheng and Gottlieb, Eli and Lu, Yiping and Cho, Kyunghyun and Wu, Jiajun and Fei-Fei, Li and Wang, Lijuan and Choi, Yejin and Li, Manling},
  journal={arXiv preprint arXiv:2504.20073},
  year={2025}
}

@article{agentevolver2025,
  title={AgentEvolver: Towards Efficient Self-Evolving Agent System},
  author={Zhai, Yunpeng and Tao, Shuchang and Chen, Cheng and Zou, Anni and Chen, Ziqian and Fu, Qingxu and Mai, Shinji and Yu, Li and Deng, Jiaji and Cao, Zouying and Liu, Zhaoyang and Ding, Bolin and Zhou, Jingren},
  journal={arXiv preprint arXiv:2511.10395},
  year={2025}
}

@inproceedings{rzero2025,
  title={{R-Zero}: Self-Evolving Reasoning {LLM} from Zero Data},
  author={Huang, Chengsong and Yu, Wenhao and Wang, Xiaoyang and Zhang, Hongming and Li, Zongxia and Li, Ruosen and Huang, Jiaxin and Mi, Haitao and Yu, Dong},
  booktitle={International Conference on Learning Representations},
  year={2026},
  url={https://openreview.net/forum?id=96apU6YzSO}
}

@article{seal2026,
  title={SEAL: Synergistic Co-Evolution of Agents and Learning Environments},
  author={Hu, Yihao and Wen, Zhihao and Liu, Xiujin and Wang, Pan and Zhang, Xin and Wu, Wei},
  journal={arXiv preprint arXiv:2605.24426},
  year={2026}
}

@inproceedings{ace2026,
  title={Agentic Context Engineering: Evolving Contexts for Self-Improving Language Models},
  author={Zhang, Qizheng and Hu, Changran and Upasani, Shubhangi and Ma, Boyuan and Hong, Fenglu and Kamanuru, Vamsidhar and Rainton, Jay and Wu, Chen and Ji, Mengmeng and Li, Hanchen and Thakker, Urmish and Zou, James and Olukotun, Kunle},
  booktitle={International Conference on Learning Representations},
  year={2026},
  url={https://openreview.net/forum?id=eC4ygDs02R}
}

@inproceedings{webevolver2025,
  title={{W}eb{E}volver: Enhancing Web Agent Self-Improvement with Co-evolving World Model},
  author={Fang, Tianqing and Zhang, Hongming and Zhang, Zhisong and Ma, Kaixin and Yu, Wenhao and Mi, Haitao and Yu, Dong},
  booktitle={Proceedings of the 2025 Conference on Empirical Methods in Natural Language Processing},
  pages={8959--8975},
  year={2025}
}

@article{worldknowledgeexploration2026,
  title={Training LLM Agents for Spontaneous, Reward-Free Self-Evolution via World Knowledge Exploration},
  author={Zhang, Qifan and Ma, Dongyang and Fang, Tianqing and Li, Jia and Tang, Jing and Chen, Nuo and Mi, Haitao and Wang, Yan},
  journal={arXiv preprint arXiv:2604.18131},
  year={2026}
}

@article{pace2026,
  title={PACE: Two-Timescale Self-Evolution for Small Language Model Agents},
  author={Ling, Chen and Chen, Pei and Guan, Albert and Qu, Jiaming and Akbar, Shayan Ali and Gopinathan, Madhu and Cornejo, Erwin},
  journal={arXiv preprint arXiv:2605.23019},
  year={2026}
}

@article{evotrainer2026,
  title={EvoTrainer: Co-Evolving LLM Policies and Training Harnesses for Autonomous Agentic Reinforcement Learning},
  author={Chen, Guhong and Shi, Yingcheng and Li, Yongbin and Li, Binhua and Xu, Xander and Wei, Hu and Ni, Shiwen and Yang, Min and Ye, Jieping},
  journal={arXiv preprint arXiv:2606.03108},
  year={2026}
}

@inproceedings{evods2026,
  title={{EvoDS}: Self-Evolving Autonomous Data Science Agent with Skill Learning and Context Management},
  author={Yang, Zherui and Liu, Fan and Ning, Yansong and Liu, Hao},
  booktitle={Proceedings of the 32nd ACM SIGKDD Conference on Knowledge Discovery and Data Mining V.2},
  pages={6128--6139},
  year={2026}
}

@article{reflexion2023,
  title={Reflexion: Language Agents with Verbal Reinforcement Learning},
  author={Shinn, Noah and Cassano, Federico and Gopinath, Ashwin and Narasimhan, Karthik and Yao, Shunyu},
  journal={Advances in Neural Information Processing Systems},
  volume={36},
  year={2023}
}

@article{textgrad2024,
  title={TextGrad: Automatic "Differentiation" via Text},
  author={Yuksekgonul, Mert and Bianchi, Federico and Boen, Joseph and Liu, Sheng and Huang, Zhi and Guestrin, Carlos and Zou, James},
  journal={arXiv preprint arXiv:2406.07496},
  year={2024}
}

@inproceedings{gepa2026,
  title={{GEPA}: Reflective Prompt Evolution Can Outperform Reinforcement Learning},
  author={Agrawal, Lakshya A. and Tan, Shangyin and Soylu, Dilara and Ziems, Noah and Khare, Rishi and Opsahl-Ong, Krista and Singhvi, Arnav and Shandilya, Herumb and Ryan, Michael J. and Jiang, Meng and Potts, Christopher and Sen, Koushik and Dimakis, Alexandros G. and Stoica, Ion and Klein, Dan and Zaharia, Matei and Khattab, Omar},
  booktitle={International Conference on Learning Representations},
  year={2026},
  url={https://openreview.net/forum?id=RQm2KQTM5r}
}

@inproceedings{dynamiccheatsheet2025,
  title={Dynamic Cheatsheet: Test-Time Learning with Adaptive Memory},
  author={Suzgun, Mirac and Yuksekgonul, Mert and Bianchi, Federico and Jurafsky, Dan and Zou, James},
  booktitle={Proceedings of the 19th Conference of the European Chapter of the Association for Computational Linguistics (Volume 1: Long Papers)},
  pages={7080--7106},
  year={2026}
}

@inproceedings{selfrag2023,
  title={Self-{RAG}: Learning to Retrieve, Generate, and Critique through Self-Reflection},
  author={Asai, Akari and Wu, Zeqiu and Wang, Yizhong and Sil, Avirup and Hajishirzi, Hannaneh},
  booktitle={International Conference on Learning Representations},
  year={2024},
  url={https://openreview.net/forum?id=hSyW5go0v8}
}

@inproceedings{bagel2024,
  title={{BAGEL}: Bootstrapping Agents by Guiding Exploration with Language},
  author={Murty, Shikhar and Manning, Christopher D. and Shaw, Peter and Joshi, Mandar and Lee, Kenton},
  booktitle={International Conference on Machine Learning},
  pages={36894--36910},
  year={2024},
  organization={PMLR}
}

@inproceedings{openwebvoyager2024,
  title={{O}pen{W}eb{V}oyager: Building Multimodal Web Agents via Iterative Real-World Exploration, Feedback and Optimization},
  author={He, Hongliang and Yao, Wenlin and Ma, Kaixin and Yu, Wenhao and Zhang, Hongming and Fang, Tianqing and Lan, Zhenzhong and Yu, Dong},
  booktitle={Proceedings of the 63rd Annual Meeting of the Association for Computational Linguistics (Volume 1: Long Papers)},
  pages={27545--27564},
  year={2025}
}

@inproceedings{godelagent2025,
  title={G{\"o}del Agent: A Self-Referential Agent Framework for Recursively Self-Improvement},
  author={Yin, Xunjian and Wang, Xinyi and Pan, Liangming and Lin, Li and Wan, Xiaojun and Wang, William Yang},
  booktitle={Proceedings of the 63rd Annual Meeting of the Association for Computational Linguistics (Volume 1: Long Papers)},
  pages={27890--27913},
  year={2025}
}

@article{agentq2024,
  title={Agent Q: Advanced Reasoning and Learning for Autonomous AI Agents},
  author={Putta, Pranav and Mills, Edmund and Garg, Naman and Motwani, Sumeet and Finn, Chelsea and Garg, Divyansh and Rafailov, Rafael},
  journal={arXiv preprint arXiv:2408.07199},
  year={2024}
}

@article{restreact2023,
  title={ReST meets ReAct: Self-Improvement for Multi-Step Reasoning LLM Agent},
  author={Aksitov, Renat and Miryoosefi, Sobhan and Li, Zonglin and Li, Daliang and Babayan, Sheila and Kopparapu, Kavya and Fisher, Zachary and Guo, Ruiqi and Prakash, Sushant and Srinivasan, Pranesh and Zaheer, Manzil and Yu, Felix and Kumar, Sanjiv},
  journal={arXiv preprint arXiv:2312.10003},
  year={2023}
}

@inproceedings{webcot2025,
  title={{W}eb{C}o{T}: Enhancing Web Agent Reasoning by Reconstructing Chain-of-Thought in Reflection, Branching, and Rollback},
  author={Hu, Minda and Fang, Tianqing and Zhang, Jianshu and Ma, Jun-Yu and Zhang, Zhisong and Zhou, Jingyan and Zhang, Hongming and Mi, Haitao and Yu, Dong and King, Irwin},
  booktitle={Findings of the Association for Computational Linguistics: EMNLP 2025},
  pages={5155--5173},
  year={2025}
}

@inproceedings{videophy2_2025,
  title={{VideoPhy-2}: A Challenging Action-Centric Physical Commonsense Evaluation in Video Generation},
  author={Bansal, Hritik and Peng, Clark and Bitton, Yonatan and Goldenberg, Roman and Grover, Aditya and Chang, Kai-Wei},
  booktitle={International Conference on Learning Representations},
  year={2026},
  url={https://openreview.net/forum?id=HA8KSQW7SO}
}

@inproceedings{phygenbench2024,
  title={Towards World Simulator: Crafting Physical Commonsense-Based Benchmark for Video Generation},
  author={Meng, Fanqing and Liao, Jiaqi and Tan, Xinyu and Lu, Quanfeng and Shao, Wenqi and Zhang, Kaipeng and Cheng, Yu and Li, Dianqi and Luo, Ping},
  booktitle={International Conference on Machine Learning},
  pages={43781--43806},
  year={2025},
  organization={PMLR}
}

@inproceedings{physicsiq2025,
  title={Do Generative Video Models Understand Physical Principles?},
  author={Motamed, Saman and Culp, Laura and Swersky, Kevin and Jaini, Priyank and Geirhos, Robert},
  booktitle={Proceedings of the IEEE/CVF Winter Conference on Applications of Computer Vision},
  pages={948--958},
  year={2026}
}

@inproceedings{paibench2025,
  title={{PAI-Bench}: A Comprehensive Benchmark For Physical {AI}},
  author={Zhou, Fengzhe and Huang, Jiannan and Li, Jialuo and Ramanan, Deva and Shi, Humphrey},
  booktitle={Proceedings of the IEEE/CVF Conference on Computer Vision and Pattern Recognition},
  pages={21522--21536},
  year={2026}
}

@article{wan2025,
  title={Wan: Open and Advanced Large-Scale Video Generative Models},
  author={Team Wan and Ang Wang and Baole Ai and Bin Wen and Chaojie Mao and Chen-Wei Xie and Di Chen and Feiwu Yu and Haiming Zhao and Jianxiao Yang and Jianyuan Zeng and Jiayu Wang and Jingfeng Zhang and Jingren Zhou and Jinkai Wang and Jixuan Chen and Kai Zhu and Kang Zhao and Keyu Yan and Lianghua Huang and Mengyang Feng and Ningyi Zhang and Pandeng Li and Pingyu Wu and Ruihang Chu and Ruili Feng and Shiwei Zhang and Siyang Sun and Tao Fang and Tianxing Wang and Tianyi Gui and Tingyu Weng and Tong Shen and Wei Lin and Wei Wang and Wei Wang and Wenmeng Zhou and Wente Wang and Wenting Shen and Wenyuan Yu and Xianzhong Shi and Xiaoming Huang and Xin Xu and Yan Kou and Yangyu Lv and Yifei Li and Yijing Liu and Yiming Wang and Yingya Zhang and Yitong Huang and Yong Li and You Wu and Yu Liu and Yulin Pan and Yun Zheng and Yuntao Hong and Yupeng Shi and Yutong Feng and Zeyinzi Jiang and Zhen Han and Zhi-Fan Wu and Ziyu Liu},
  journal={arXiv preprint arXiv:2503.20314},
  year={2025}
}

@article{nvidia2026cosmos3omnimodalworld,
  title={Cosmos 3: Omnimodal World Models for Physical AI},
  author={NVIDIA and Aditi and Niket Agarwal and Arslan Ali and Jon Allen and Martin Antolini and Adeline Aubame and Alisson Azzolini and Junjie Bai and Maciej Bala and Yogesh Balaji and Josh Bapst and Aarti Basant and Mukesh Beladiya and Mohammad Qazim Bhat and Zaid Pervaiz Bhat and Dan Blick and Vanni Brighella and Han Cai and Tiffany Cai and Eric Cameracci and Jiaxin Cao and Yulong Cao and Mark Carlson and Carlos Casanova and Ting-Yun Chang and Yan Chang and Yu-Wei Chao and Prithvijit Chattopadhyay and Roshan Chaudhari and Chieh-Yun Chen and Junyu Chen and Ke Chen and Qizhi Chen and Wenkai Chen and Xiaotong Chen and Yu Chen and An-Chieh Cheng and Click Cheng and Xiu Chia and Jeana Choi and Chaeyeon Chung and Wenyan Cong and Yin Cui and Magdalena Dadela and Nalin Dadhich and Wenliang Dai and Joyjit Daw and Alperen Degirmenci and Rodrigo Vieira Del Monte and Robert Denomme and Sameer Dharur and Marco Di Lucca and Ke Ding and Wenhao Ding and Yifan Ding and Yuzhu Dong and Nicole Drumheller and Yilun Du and Aigul Dzhumamuratova and Aleksandr Efitorov and Hamid Eghbalzadeh and Naomi Eigbe and Imad El Hanafi and Hassan Eslami and Benedikt Falk and Jiaojiao Fan and Jim Fan and Amol Fasale and Sergiy Fefilatyev and Liang Feng and Francesco Ferroni and Sanja Fidler and Xiao Fu and Vikram Fugro and Prashant Gaikwad and TJ Galda and Katelyn Gao and Yihuai Gao and Wenhang Ge and Sreyan Ghosh and Arushi Goel and Vivek Goel and Akash Gokul and Rama Govindaraju and Jinwei Gu and Miguel Guerrero and Elfie Guo and Aryaman Gupta and Siddharth Gururani and Hugo Hadfield and Song Han and Ankur Handa and Zekun Hao and Mohammad Harrim and Ali Hassani and Nathan Hayes-Roth and Yufan He and Chris Helvig and Cyrus Hogg and Madison Huang and Michael Huang and Sophia Huang and Yufan Huang and Jacob Huffman and DeLesley Hutchins and Suneel Indupuru and Boris Ivanovic and Arihant Jain and Joel Jang and Ryan Ji and Yanan Jian and Dongfu Jiang and Jingyi Jin and Atharva Joshi and Nikhilesh Joshi and Pranjali Joshi and Andy Ju and Jaehun Jung and Weiwei Kang and Scott Kassekert and Jan Kautz and Ashna Khetan and Julia Kiczka and Slawek Kierat and Gwanghyun Kim and Kuno Kim and Sunny Kim and Kezhi Kong and Xin Kong and Zhifeng Kong and Tomasz Kornuta and Egor Krivov and Hui Kuang and Saurav Kumar and Chia-Wen Kuo and George Kurian and Wojciech Kutak and JF Lafleche and Himangshu Lahkar and Omar Laymoun and Jayjun Lee and Sanggil Lee and Gabriele Leone and Boyi Li and Freya Li and Jiajun Li and Jinfeng Li and Ling Li and Pengcheng Li and Shangru Li and Tingle Li and Xiaolong Li and Xuan Li and Zhaoshuo Li and Zhiqi Li and Hao Liang and Maosheng Liao and Chen-Hsuan Lin and Tsung-Yi Lin and Ming-Yu Liu and Sifei Liu and Zihan Liu and Hai Loc Lu and Xiangyu Lu and Alice Luo and Ruipu Luo and Wenjie Luo and Jiangran Lyu and Martin Ding Ma and Nic Ma and Qianli Ma and Dawid Majchrowski and Louis Marcoux and Miguel Martin and Qing Miao and Ashkan Mirzaei and Shreyas Misra and Kaichun Mo and Durra Mohsin and Hyejin Moon and Pawel Morkisz and Saeid Motiian and Kirill Motkov and Seungjun Nah and Yashraj Narang and Deepak Narayanan and Thabang Ngazimbi and Julian Ouyang and Shubham Pachori and David Page and Yatian Pang and Sehwi Park and Mahesh Patekar and Mostofa Patwary and Marco Pavone and Trung Pham and Wei Ping and Soha Pouya and Shrimai Prabhumoye and Varun Praveen and Delin Qu and Hesam Rabeti and Morteza Ramezanali and Marilyn Reeb and Xuanchi Ren and Kristen Rumley and Wojciech Rymer and Jun Saito and Yeongho Seol and John Shao and Piyush Shekdar and Tianwei Shen and Humphrey Shi and Min Shi and Stella Shi and Kevin Shih and Mohammad Shoeybi and Mateusz Sieniawski and Shuran Song and Alexander Sotelo and Amir Sotoodeh and Sunil Srinivasa and Vignesh Srinivasakumar and Bartosz Stefaniak and Rahul Heinrich Steiger and Shangkun Sun and Jiaxiang Tang and Shitao Tang and Yangyang Tang and Yue Tang and Tolou Tavakkoli and Kayley Ting and Krzysztof Tomala and Wei-Cheng Tseng and Jibin Varghese and Sergei Vasilev and Thomas Volk and Raju Wagwani and Roger Waleffe and Andrew Z. Wang and Boxiang Wang and Haoxiang Wang and Qiao Wang and Shihao Wang and Shijie Wang and Ting-Chun Wang and Yan Wang and Yu Wang and Rohit Watve and David Wehr and Fangyin Wei and Xinshuo Weng and Jay Zhangjie Wu and Kedi Wu and Hongchi Xia and Summer Xiao and Tianjun Xiao and Kevin Xie and Daguang Xu and Jiashu Xu and Mengyao Xu and Ruqing Xu and Xingqian Xu and Yao Xu and Dinghao Yang and Dong Yang and Hans Yang and Xiaodong Yang and Xuning Yang and Yichu Yang and Yurong You and Zhiding Yu and Hao Yuan and Simon Yuen and Xiaohui Zeng and Pengcuo Zeren and Cindy Zha and Haotian Zhang and Jenny Zhang and Jing Zhang and Liangkai Zhang and Paris Zhang and Shun Zhang and Xuanmeng Zhang and Zhizheng Zhang and Ann Zhao and Yilin Zhao and Yuliya Zhautouskaya and Charles Zhou and Fengzhe Zhou and Shilin Zhu and Yuke Zhu and Dima Zhylko and Artur Zolkowski},
  journal={arXiv preprint arXiv:2606.02800},
  year={2026}
}

@inproceedings{yang2024cogvideox,
  title={{CogVideoX}: Text-to-Video Diffusion Models with An Expert Transformer},
  author={Yang, Zhuoyi and Teng, Jiayan and Zheng, Wendi and Ding, Ming and Huang, Shiyu and Xu, Jiazheng and Yang, Yuanming and Hong, Wenyi and Zhang, Xiaohan and Feng, Guanyu and others},
  booktitle={International Conference on Learning Representations},
  year={2025},
  url={https://openreview.net/forum?id=LQzN6TRFg9}
}

@article{hunyuanvideo2025,
  title={HunyuanVideo 1.5 Technical Report},
  author={{Tencent Hunyuan Foundation Model Team}},
  journal={arXiv preprint arXiv:2511.18870},
  year={2025}
}

@inproceedings{phygdpo,
  title={{PhyGDPO}: Physics-Aware Groupwise Direct Preference Optimization for Physically Consistent Text-to-Video Generation},
  author={Cai, Yuanhao and Li, Kunpeng and Jia, Menglin and Wang, Jialiang and Sun, Junzhe and Liang, Feng and Chen, Weifeng and Juefei-Xu, Felix and Wang, Chu and Thabet, Ali and Dai, Xiaoliang and Ju, Xuan and Yuille, Alan and Hou, Ji},
  booktitle={European Conference on Computer Vision},
  pages={91--109},
  year={2026},
  organization={Springer}
}

@article{liu2025bootstrappingphysicsgroundedvideogeneration,
  title={Bootstrapping Physics-Grounded Video Generation through VLM-Guided Iterative Self-Refinement},
  author={Liu, Yang and Zhao, Xilin and Wen, Peisong and Dai, Siran and Huang, Qingming},
  journal={arXiv preprint arXiv:2511.20280},
  year={2025}
}

@misc{kandinskywm2026,
  title={Kandinsky WM 1.0: A Family of Models for Physical AI},
  author={{Kandinsky Lab}},
  howpublished={\url{https://github.com/kandinskylab/kandinsky-wm}},
  year={2026},
  note={Image-to-video models for autonomous driving, robotics, and general physics}
}

@inproceedings{chung2023unimaxfairereffectivelanguage,
  title={{UniMax}: Fairer and More Effective Language Sampling for Large-Scale Multilingual Pretraining},
  author={Chung, Hyung Won and Constant, Noah and Garcia, Xavier and Roberts, Adam and Tay, Yi and Narang, Sharan and Firat, Orhan},
  booktitle={International Conference on Learning Representations},
  year={2023},
  url={https://openreview.net/forum?id=kXwdL1cWOAi}
}

@article{shang2026phizero,
  title={PhiZero: A World Model Built Around Physical Language},
  author={Shang, Shuyao and Wang, Yuqi and Gao, Ruopeng and Chen, Xu and Tan, Tieniu and Fan, Lue and Zhang, Zhaoxiang},
  journal={arXiv preprint arXiv:2607.28624},
  year={2026}
}

@article{lin2026phygroundbenchmarkingphysicalreasoning,
  title={PhyGround: Benchmarking Physical Reasoning in Generative World Models},
  author={Lin, Juyi and Akbari, Arash and He, Yumei and Zhao, Lin and Zhang, Haichao and Akbari, Arman and Xu, Xingchen and Lu, Zoe Y. and Nan, Enfu and Deng, Hokin and Yeh, Edmund and Ostadabbas, Sarah and Fu, Yun and Dy, Jennifer and Zhao, Pu and Wang, Yanzhi},
  journal={arXiv preprint arXiv:2605.10806},
  year={2026}
}

@misc{rädsch2026physicsiqverified,
      title={Physics-IQ Verified}, 
      author={Tim Rädsch and Yuki M Asano and Hilde Kuehne and Stefan Bauer and Priyank Jaini and Robert Geirhos and Carsten T. Lüth},
      year={2026},
      eprint={2606.18943},
      archivePrefix={arXiv},
      primaryClass={cs.CV},
      url={https://arxiv.org/abs/2606.18943}, 
}

@misc{wang2025wisa,
                title={WISA: World Simulator Assistant for Physics-Aware Text-to-Video Generation}, 
                author={Jing Wang and Ao Ma and Ke Cao and Jun Zheng and Zhanjie Zhang and Jiasong Feng and Shanyuan Liu and Yuhang Ma and Bo Cheng and Dawei Leng and Yuhui Yin and Xiaodan Liang},
                year={2025},
                eprint={2502.08153},
                archivePrefix={arXiv},
                primaryClass={cs.CV},
                url={https://arxiv.org/abs/2502.08153}, 
}

@inproceedings{xue2025phyt2v,
  title     = {{PhyT2V}: {LLM}-Guided Iterative Self-Refinement for Physics-Grounded Text-to-Video Generation},
  author    = {Xue, Qiyao and Yin, Xiangyu and Yang, Boyuan and Gao, Wei},
  booktitle = {Proceedings of the IEEE/CVF Conference on Computer Vision and Pattern Recognition (CVPR)},
  pages     = {18826--18836},
  year      = {2025}
}
}

\clearpage

\appendix
\section*{Appendix}

\section{Prompt Self-Evolution Analysis}
\label{sec}
\label{sec:loop_full}

\begin{figure}[H]
\centering
\includegraphics[width=\textwidth]
{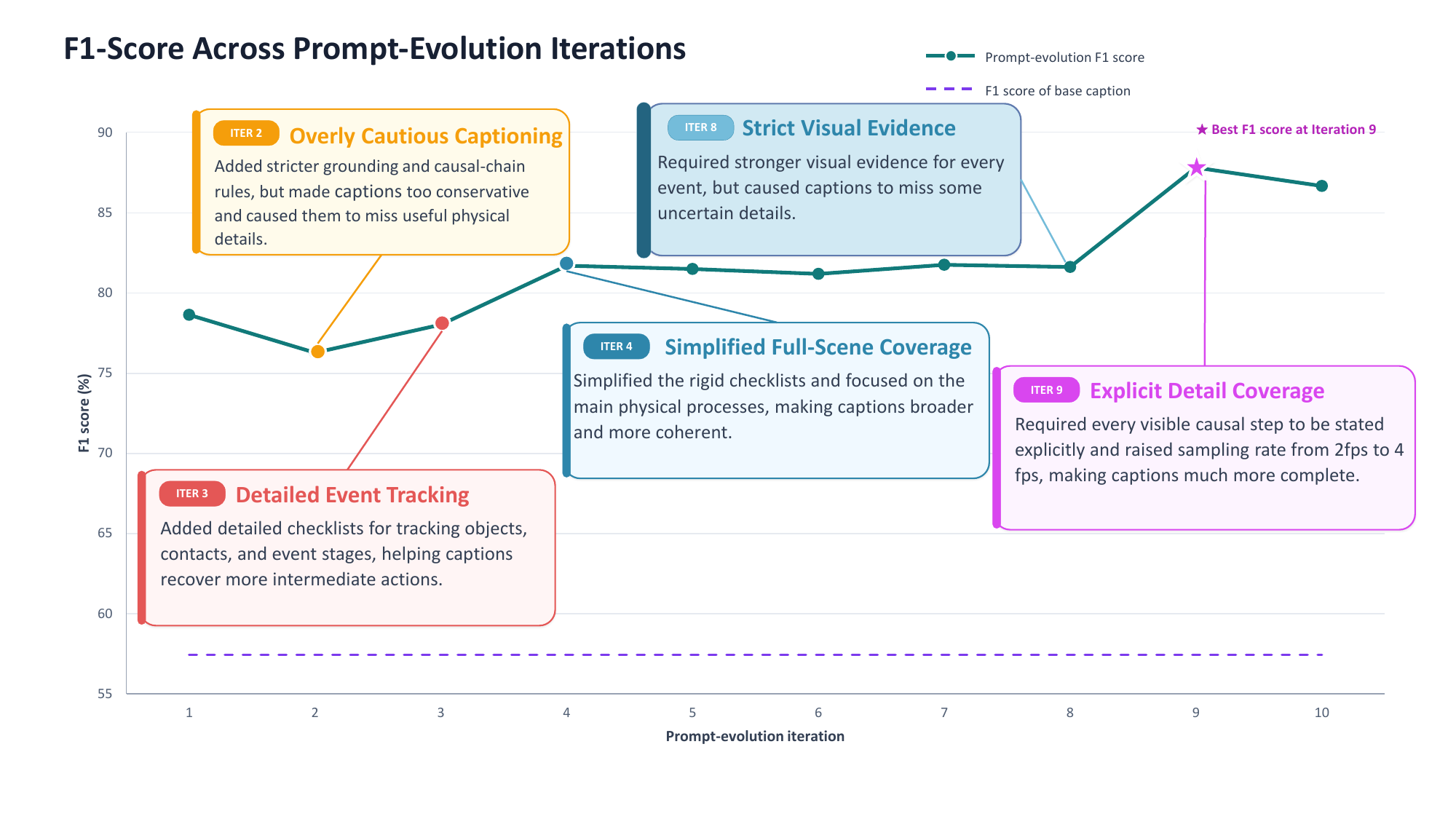}
\caption{
\textbf{F1-score trajectory across prompt self-evolution iterations on PhysCapBench}. The dashed purple line denotes the F1 score of base caption. The five callouts summarize the representative prompt
modifications introduced at Iterations~2, 3, 4, 8  and~9. The star identifies
the best-performing iteration. The annotations describe the prompt-design
changes, whereas all plotted F1 scores are obtained from PhysCapBench.
}
\label{fig}
\end{figure}

As mentioned in Sec.~\ref{sec:loop_effectiveness}, to better understand how the captioning prompt evolves during iterations, here we visualize the F1-score curve on \textbf{PhysCapBench} in Fig.~\ref{fig}. The annotations summarize five observed representative modifications to the prompt design.

\clearpage

\section{Additional Implementation Details}
\label{sec:appendix_implementation_details}
\subsection{Physcapbench and Evaluation Protocol}
\label{app:physics-caption-benchmark}
As illustrated in Sec.~\ref{sec:physcapbench}, our PhysCapBench contains 246 videos and includes 3,794 human-verified physical-dynamics assertions with a mean of 15.42 assertions per video. All captions are evaluated by Gemini-3-Flash, using temperature 0.1, a 65,536-token JSON-output limit, and batches of at most 30 claims. Precision decomposes the caption into atomic claims, verifies each against the video, and computes $N_{\mathrm{correct}}/(N_{\mathrm{correct}}+N_{\mathrm{incorrect}})$. Recall denotes the proportion of the 3,794 reference assertions supported by the generated caption.

\subsection{More Details on the Agentic Loop}
\label{app:loop_detail}
Here we elaborate more implementation details for our agentic loop described in Sec.~\ref{sec:prompt_description}. We optimized the physics captions through a generate–critic–revise loop on a fixed development set of 20 videos, which is randomly sampled from our training set, with 273 human-verified atomic physics assertions in total. At each iteration, a fixed GPT-5.5 captioner processed temporally ordered frames and jointly generated \textit{physics\_reasoning} and \textit{physics\_negative\_prompt} in a single call, while only \textit{physics\_reasoning} was scored. We use Gemini-3.1-Pro to measure assertion-level recall and video-grounded claim precision, and the evolution agent used both aggregate scores and claim-level failure rationales to revise the prompt. 

To construct the upsampling prompt, we preserved the physics-focused requirements of the final caption prompt while adapting its input interface. Instead of describing the visual content from scratch, the upsampler expands the original short benchmark caption, together with visual inputs when required, into the same structured \textit{physics\_reasoning} and \textit{physics\_negative\_prompt} fields.

\subsection{Long Context Support on Wan}
\label{app:wan_longcontext_detail}
\begin{figure}[h]
    \centering
    \includegraphics[width=0.8\linewidth]{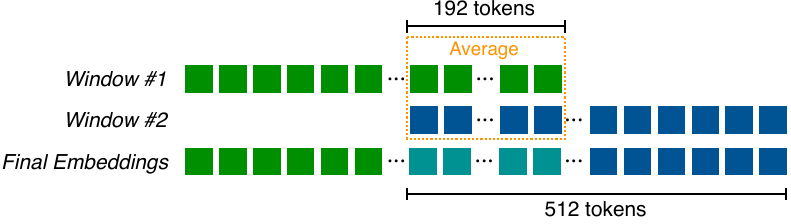}
    \caption{\textbf{Sliding-window mechanism for text encoding in Wan2.1.}
    }
    \label{fig:sliding_window}
\end{figure}

In Sec.~\ref{sec:experimental_details}, one of the backbones going through our fine-tuning is Wan2.1~\citep{wan2025}. The text encoder UMT5-XXL~\citep{chung2023unimaxfairereffectivelanguage} of Wan2.1 imposes a maximum context length of 512 tokens, which can be restrictive for our physics-aware prompts that contain both detailed descriptions and reasoning. To accommodate arbitrarily long captions without modifying the pretrained text encoder, we adopt a sliding-window strategy. Specifically, an over-length prompt is partitioned into overlapping windows, each within the 512-token limit, and the resulting text representations are aggregated to provide conditioning for the video generation model. In practice, the overlap between two consecutive context windows is set to 192 tokens, and we take the average of the embeddings from the two windows over the overlapping region, as shown in Fig.~\ref{fig:sliding_window}.

\subsection{More Details on Physics Video Generation Benchmarks}
\label{app:benchmark_detail}
In Sec.~\ref{sec:experiment}, we evaluate models on four benchmarks: PhyGenBench and VideoPhy-2 for Text-to-Video generation, and Physics-IQ Verified and PhyGround for Image-to-Video generation. In this section, we briefly describe the evaluation protocol and the physical capabilities assessed by each benchmark.

\paragraph{PhyGenBench.}
The 160 prompts of PhyGenBench can be divided into four categories of physical phenomena: \textit{force}, \textit{light}, \textit{heat}, and \textit{material}. We calculate the video-level score with the official scripts, but replacing the evaluator with GPT-5.5. Each video is graded on a 0--3 scale, where 0 indicates a complete fantastical physical behavior and 3 indicates strong physical realism. The final score is obtained by averaging across all 160 samples.

\paragraph{VideoPhy-2.}
VideoPhy-2 evaluates generated videos from two aspects: \textit{semantic consistency} (SA) and \textit{physical correctness} (PC), each is graded on a 0--5 scale. Its test set contains 591 samples covering common physical phenomena in everyday scenarios, with 178 challenging samples selected to form a Hard subset. We replace its offline VLM evaluator with GPT-5.5 (App.~\ref{sec:gpt55}). The joint accuracy measures the fraction of videos that simultaneously satisfy both $\text{SA}\geq 4$ and $\text{PC}\geq 4$.

\paragraph{Physics-IQ Verified.}
Physics-IQ (verified version) evaluates generated videos across multiple physical concepts using four complementary metrics: \textit{spatial consistency} (S), \textit{spatial-temporal consistency} (ST), \textit{weighted-spatial consistency} (WS), and \textit{mean squared error} (MSE). It contains 66 physical scenarios, with three perspectives evaluated per scenario (198 samples in total). It provides ground-truth videos for reference. The final score is computed by aggregating the scores across these metrics and averaging over all evaluated samples.

\paragraph{PhyGround.}
PhyGround evaluates each generated video using \textit{semantic adherence} (General) and \textit{physical temporal validity} (Physics). Each criterion is scored on a 1--5 scale, and averaged across all 250 samples.

\subsection{Training Details}
Here we offer training details for Cosmos3, Wan2.1 and our PhysThinker in Sec.~\ref{sec:experiment}.
\subsubsection{Cosmos~3 family}
\label{app:cosmos_train_detail}
\paragraph{Cosmos3-Nano.}
We fine-tuned Cosmos3-Nano using the dataset of about 183K video–caption pairs (Sec.~\ref{sec:experimental_details}), including 71K WISA-80K samples retained after filtering for motion and aesthetic quality and removing multi-shot or metadata-incomplete videos, plus 112K videos selected through our deficiency-guided retrieval data pipeline. Each video was paired with a caption combining the base visual description and the generated physics-reasoning annotation. Videos were processed at a spatial resolution of ($256\times256$), with a maximum caption length of 4,096 tokens and a maximum packed sequence length of 45,056 tokens. During training, the conditioning format was sampled as 70\% text-to-video, 20\% image-to-video using the first frame, and 10\% video-to-video using the first five frames; classifier-free guidance dropout was set to 0.1. We applied LoRA only to the query, key, value, and output projections of the attention layers, using rank 16 and scaling factor 32. Training used bfloat16 precision, fully sharded data parallelism, and full activation checkpointing across 16 GPUs on two nodes. Each GPU processed one packed sequence at a time, with four gradient-accumulation steps, giving a nominal global batch size of 64 packed sequences per optimizer update. We optimized the model with fused AdamW using a learning rate of ($5\times10^{-4}$), ($\beta_1=0.9$), ($\beta_2=0.95$), ($\epsilon=10^{-6}$), no weight decay, and gradient clipping at 0.1. The checkpoint at iteration 1,340 was used for the reported benchmark evaluation.

\paragraph{Cosmos3-Super.}
We fine-tuned Cosmos3-Super on the same training set and
caption format described above. Videos were resized and
center-cropped using aspect-ratio-dependent buckets at the
256-resolution tier (e.g., $256\times256$ for square videos
and $320\times192$ for landscape 16:9 videos), with a maximum
caption length of 4,096 tokens and a maximum packed sequence
length of 45,056 tokens. The conditioning format was sampled
as 70\% text-to-video, 20\% image-to-video using the first
frame, and 10\% video-to-video using the first two latent
frames, corresponding to five video frames under temporal
compression by a factor of four. Classifier-free guidance
dropout was set to 0.1. We applied LoRA only to the query,
key, value, and output projections of the generation
expert's attention layers, using rank 16 and scaling
factor 32. Training used bfloat16 precision, fully sharded
data parallelism, and full activation checkpointing across
16 GPUs on four nodes, with context-parallel degree 4
and data-parallel degree 4. Each data-parallel replica
processed one packed sequence per microstep, distributed
across its four context-parallel ranks. With two
gradient-accumulation steps, this yielded a nominal global
batch size of eight packed sequences per optimizer update.
We used fused AdamW with a learning rate of
$5\times10^{-4}$, $\beta_1=0.9$, $\beta_2=0.95$,
$\epsilon=10^{-6}$, zero weight decay, and gradient clipping
at 0.1. The checkpoint at iteration 1000 was used for evaluation.

\paragraph{Cosmos3-Edge.}
We fine-tuned Cosmos3-Edge from its pretrained initialization
on the same training set and caption format described above.
Videos were processed using aspect-ratio-dependent buckets
at the 480-resolution tier (e.g., $832\times480$ for landscape
16:9 videos), with at most 121 frames per video and a maximum
caption length of 8,192 tokens. Training used sample-count-based
packing with at most two samples per packed batch and a
model token capacity of 90,112. The conditioning format
was sampled as 70\% text-to-video and 30\% image-to-video
using the first frame; video-to-video conditioning was not
used. Classifier-free guidance dropout was set to 0.1.
We applied LoRA only to the query, key, value, and output
projections of the generation expert's attention layers,
using rank 16 and scaling factor 32. The main training run
used bfloat16 precision, fully sharded data parallelism,
and full activation checkpointing across 32 GPUs on eight
nodes, with context-parallel degree 1, data-parallel degree
32, and no gradient accumulation. This corresponds to a
nominal global batch size of 64 samples per optimizer
update when every packed batch contains two samples.
We used fused AdamW with a learning rate of $5\times10^{-4}$,
$\beta_1=0.9$, $\beta_2=0.95$, $\epsilon=10^{-6}$, zero
weight decay, and gradient clipping at 0.1. Checkpoints were
evaluated at 500-update intervals; the checkpoint at
iteration 2,000 was used for the reported benchmark evaluation.

\subsubsection{Wan2.1}
\label{app:wan_train_detail}
We fine-tuned both Wan2.1-T2V-14B and Wan2.1-I2V-14B using the identical dataset described in~\ref{app:cosmos_train_detail}. Each caption contains both the base visual description and the generated physics-reasoning annotation. During training, classifier-free guidance dropout was set to 0.1. We applied LoRA only to the query, key, value, and output projections of the generation expert’s attention layers, using rank 256 and scaling factor 512, while the patch embedder and output modules are fully fine-tuned. Training used bfloat16 precision, fully sharded data parallelism, and full activation checkpointing across 16 NVIDIA H100 GPUs. The global batch size is set to 32. We optimized the model with fused AdamW using a learning rate of ($5\times10^{-4}$) with 1000 warmup steps, ($\beta_1=0.9$), ($\beta_2=0.95$), ($\epsilon=10^{-6}$), no weight decay, and gradient clipping at 0.1. The checkpoint at iteration 8,000 was used for the reported benchmark evaluation.

\subsubsection{PhysThinker}
\label{app:physthinker_train_detail}

PhysThinker consists of two Qwen3-VL-4B-Instruct models fine-tuned with our training data, differing only in the conditioning modality. The captioner is trained on videos paired with GPT-generated physics captions, while the upsampler is trained on short text prompts, with or without a conditioning image, paired with GPT-expanded physics-rich prompts. We applied LoRA with rank 8, scaling factor 32, and dropout 0.05 to all linear projections of the language model, while the entire vision stack (ViT and aligner) is fully fine-tuned. Training used bfloat16 precision, FlashAttention-2, DeepSpeed ZeRO-3 with optimizer and parameter offload to host memory, and full activation checkpointing including the vision tower. Examples are packed to a 57,344-token budget, with video sampled at 4 fps up to 64 frames. The global batch size is 256 packed sequences. We optimized with fused AdamW at a learning rate of ($1\times10^{-4}$) for the language-model adapters and ($2\times10^{-5}$) for the vision tower and aligner, using a cosine schedule with 5\% linear warmup, ($\beta_1=0.9$), ($\beta_2=0.95$), ($\epsilon=10^{-8}$), weight decay 0.1, and gradient clipping at 1.0. Both models were trained for 3 epochs.

The video-conditioned reasoner takes a video alone and was trained across 128 A100 GPUs. The checkpoint at step 650 was used for the reported benchmark evaluation. The prompt upsampler expands a short generation prompt, and is trained as a two-mode mixture: text-only, and text together with the clip's first frame. Each corpus contributes to both modes, trained across 32 A100 GPUs. The final checkpoint at step 258 was used for the reported benchmark evaluation.

\clearpage

\section{More Qualitative Comparisons}
\label{sec:appendix_qualitative_comparison}
As a supplement to Sec.~\ref{sec:sota_comparison}, here we show the qualitative examples across the evaluated benchmarks.

\begin{figure}[H]
  \centering
  \includegraphics[width=\textwidth,height=0.88\textheight,keepaspectratio]{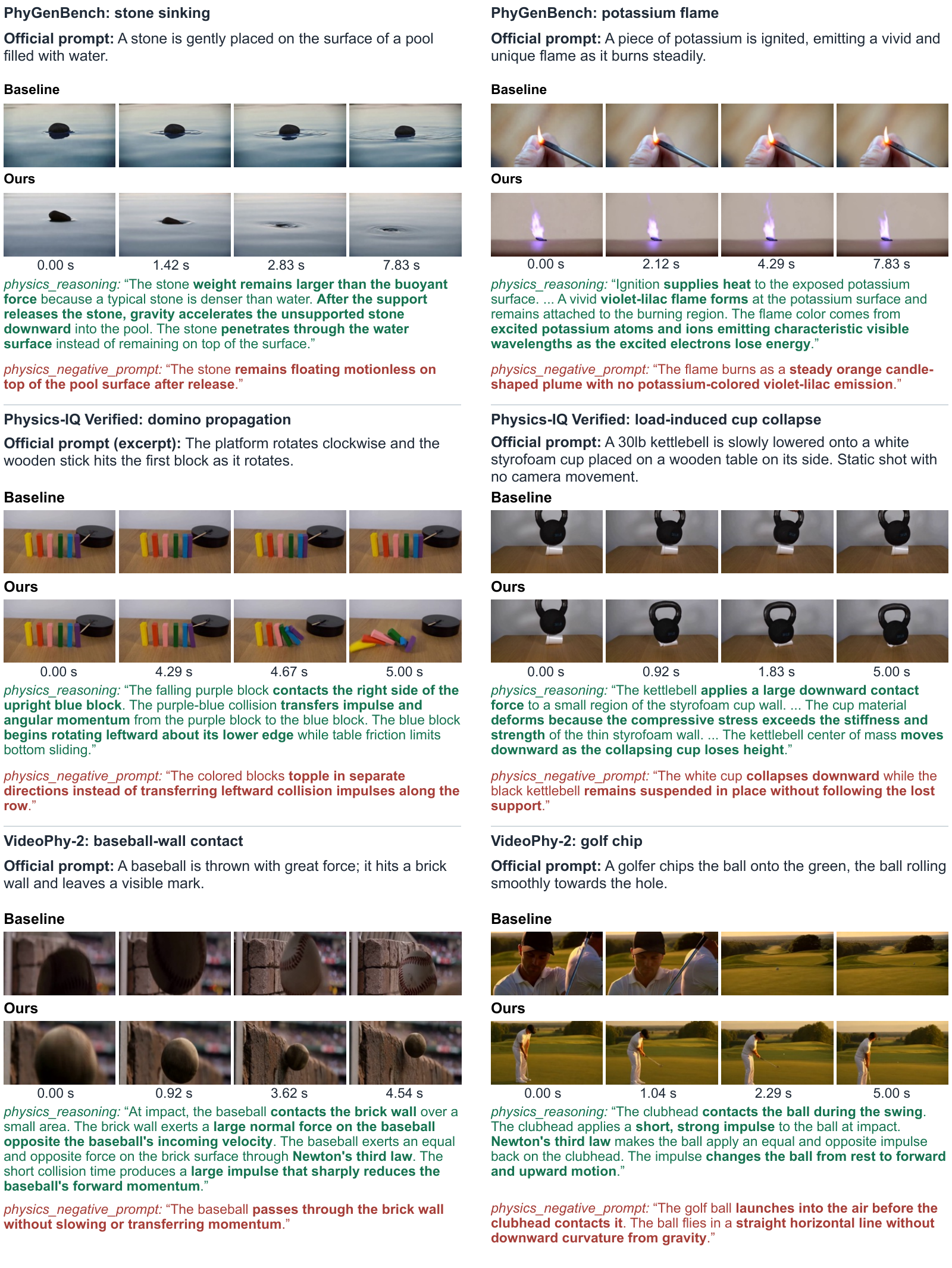}
  \caption{\textbf{Qualitative comparisons on different physics benchmarks.}
Each case compares frames generated by the Cosmos3-Nano base model using base caption (Baseline) with those generated by our trained model using base caption augmented with our \textit{physics\_reasoning} and \textit{physics\_negative\_prompt} (Ours) at identical timestamps. Official prompts
and selected physics reasoning and negative-prompt excerpts are shown,
with bold phrases highlighting physical constraints. Negative excerpts
describe violations to avoid. All frames are uncropped.}
  \label{fig:more-physics-combined}
\end{figure}
\clearpage

\section{Demos in Driving and Robotics Scenarios}
As stated in Sec.~\ref{sec:sota_comparison}, here we offer 6 demos to illustrate the strong performance of our method in driving and robotics scenarios.
\label{sec:demo_driving_robotics}

\begin{figure}[H]
	\centering
	\includegraphics[width=\textwidth,height=0.88\textheight,keepaspectratio]{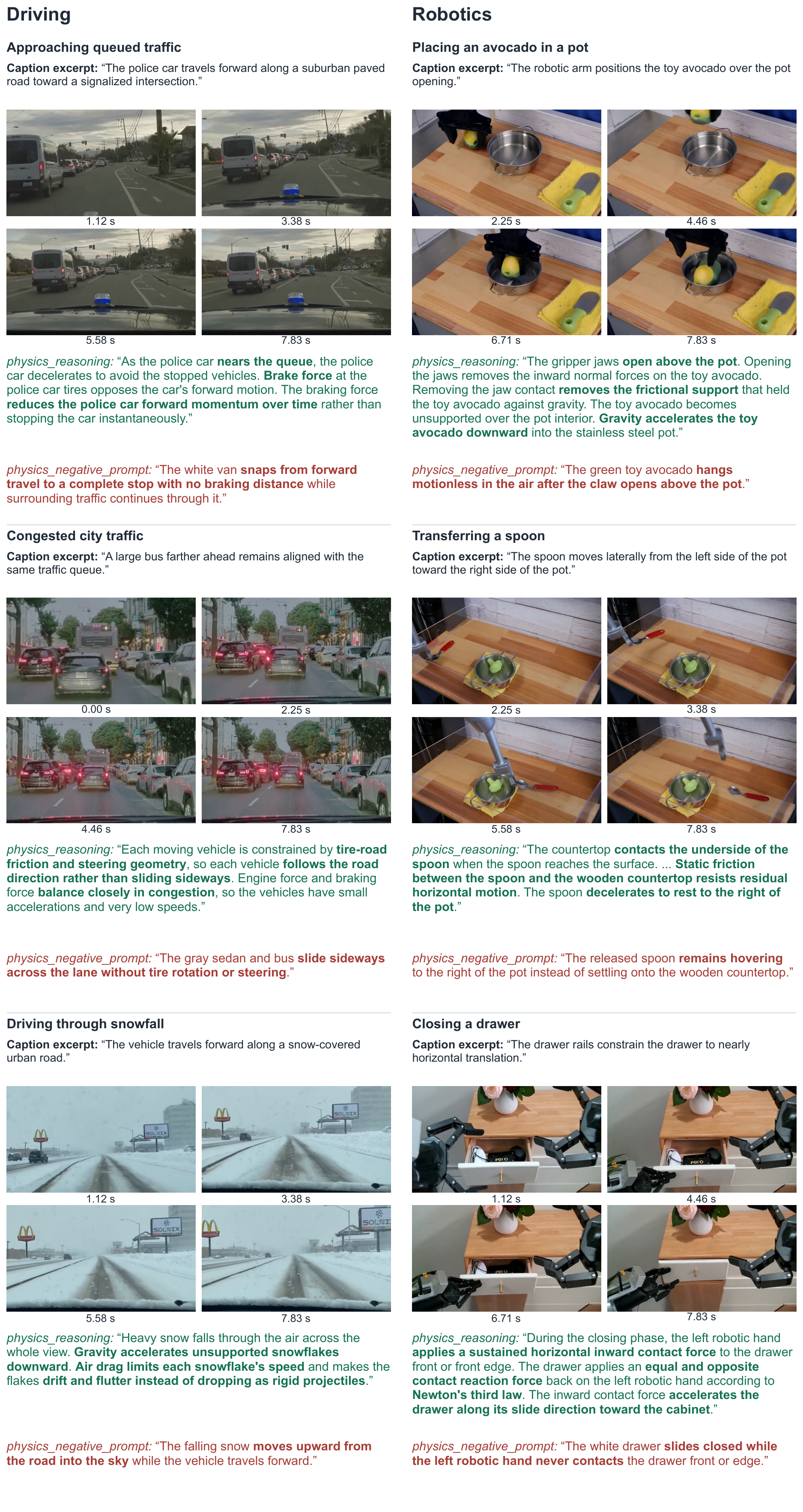}
	\caption{\textbf{Qualitative results on driving and robotics.}
	All sequences are generated by our trained model using base caption augmented with our \textit{physics\_reasoning} and \textit{physics\_negative\_prompt}. Each example includes four uncropped video
	frames with timestamps, a caption excerpt, and selected physics reasoning
	and negative-prompt excerpts. }
	\label{fig:paibench-driving-robotics}
\end{figure}

\clearpage

\section{More Examples of PhysCapBench}
\label{sec:examples_physcapbench}
We provide two examples from PhysCapBench (Sec.~\ref{sec:physcapbench}): a falling cup that fractures upon impact and a Newton's cradle exhibiting coupled collisions. Each example includes six chronological video frames and its complete set of 20 human-curated annotations, categorized as \textit{Cause}, \textit{Law}, or \textit{Effect}. Together, these examples illustrate how the benchmark captures initiating interactions, governing physical principles, and observable outcomes beyond coarse event descriptions.

\begin{figure}[H]
    \centering
    \includegraphics[
        page=1,
        trim=30mm 70mm 35mm 50mm,
        clip,
        width=\linewidth,
        height=0.78\textheight,
        keepaspectratio
    ]{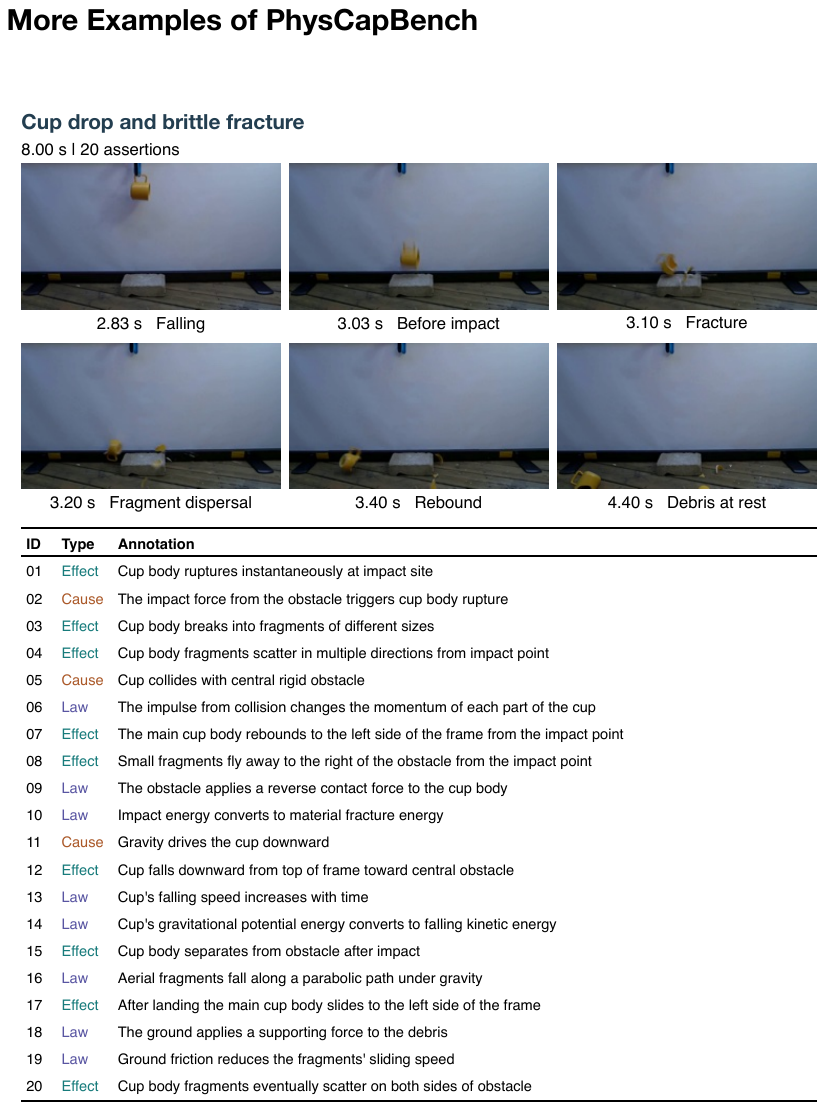}
    \caption{
    \textbf{Cup drop, impact, and fracture.} Six chronological frames accompany
    the reference physical assertions in their original wording and rank order.
    Phase labels summarize the displayed frames; assertion IDs are not temporal alignments.
    }
    \label{fig:physcapbench_cup}
\end{figure}

\clearpage

\begin{figure}[H]
    \centering
    \includegraphics[
        page=2,
        trim=32mm 75mm 32mm 37mm,
        clip,
        width=\linewidth,
        height=0.86\textheight,
        keepaspectratio
    ]{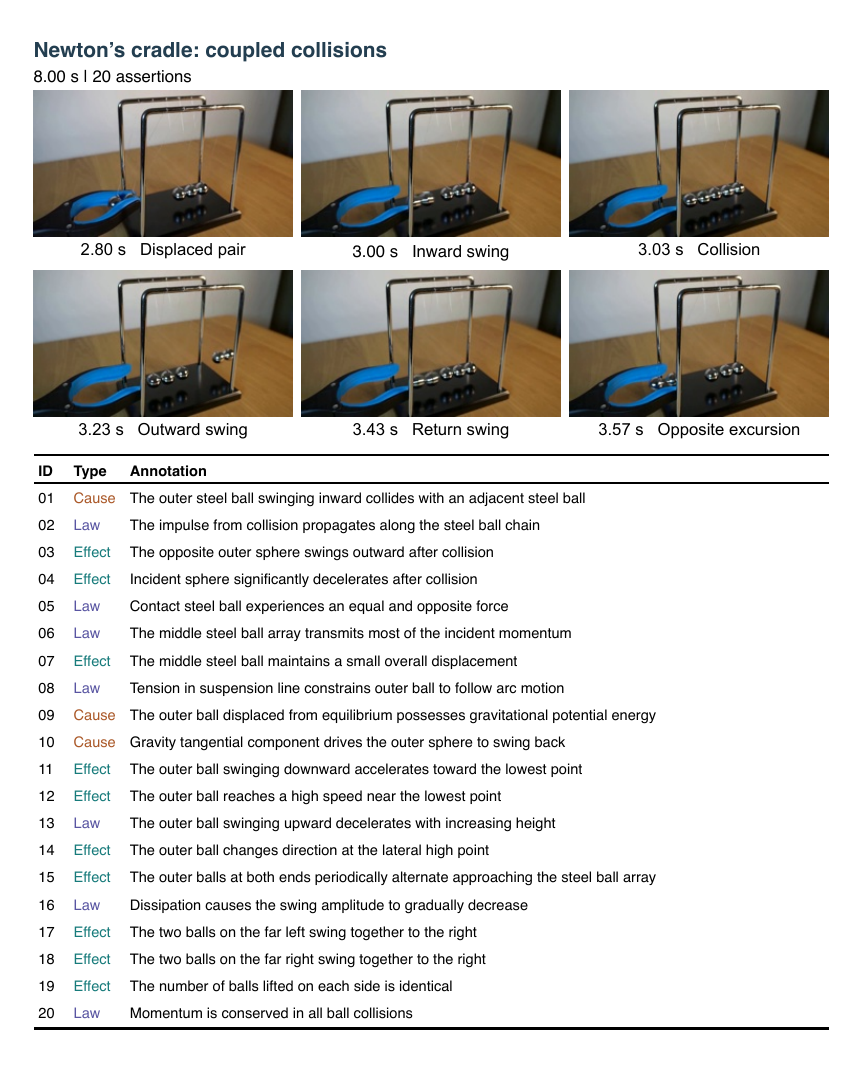}
    \caption{
    \textbf{Newton's cradle and momentum transfer.} Six chronological frames accompany
    the reference physical assertions in their original wording and rank order.
    Phase labels summarize the displayed frames; assertion IDs are not temporal alignments.
    }
    \label{fig:physcapbench_newton_cradle}
\end{figure}

\clearpage

\section{Analysis on Video Generation Attention Maps}
\label{sec:analysis_attention}
As in Sec.~\ref{sec:loop_effectiveness}, to measure the effect of \textit{physics\_reasoning} derived by self-evolution iterations in Physis-Lang, we use Cosmos3-Nano as the backbone and compare two configurations: inference with base caption only using the pretrained model, and inference with base caption + \textit{physics\_reasoning} using the model after SFT on the data we re-captioned with \textit{physics\_reasoning} field. 
Cosmos3 processes text tokens and video latent tokens in a unified attention
sequence. To measure the importance of each token, we compute the attention from sampled
visual queries to all text tokens. Specifically, we capture the normalized
query and key features after mRoPE and compute the text-conditional attention
\begin{equation}
A(w, q)=\frac{1}{H}\sum_{h=1}^{H}
\operatorname{Softmax}_{w'}\!\left(
\frac{Q_h(q)K_h(w')^\top}{\sqrt{d_h}}
\right)_{w},
\end{equation}
where $w$ is the given text token, and $q$ is the video-latent query. $H$ denotes the number
of attention heads. The score of a token is obtained by averaging $A(w,q)$
over the uniformly sampled visual tokens, denoising steps, and all transformer
layers. For better visualization, we divide each word score by the mean score of all
word occurrences in the same prompt. Therefore, the larger the scaled score of a text token is, the greater the role that corresponding word plays in video generation.

Similarly, to obtain the focal regions of an RGB frame with respect to a key sentence $S$, we average the attention assigned to all text tokens within the span of $S$:
\begin{equation}
M_S(q)=\frac{1}{|\mathcal{T}(S)|}
\sum_{w\in\mathcal{T}(S)}A(w,q),
\end{equation}
where $q$ is a visual latent token encoded from the RGB frame, and $\mathcal{T}(S)$ denotes the corresponding token set. Each sampled
visual query retains its latent coordinate $(t,y,x)$, allowing the values
$M_S(q)$ to be arranged into a spatial map.

\begin{figure}[h]
    \centering
    \includegraphics[width=\linewidth]{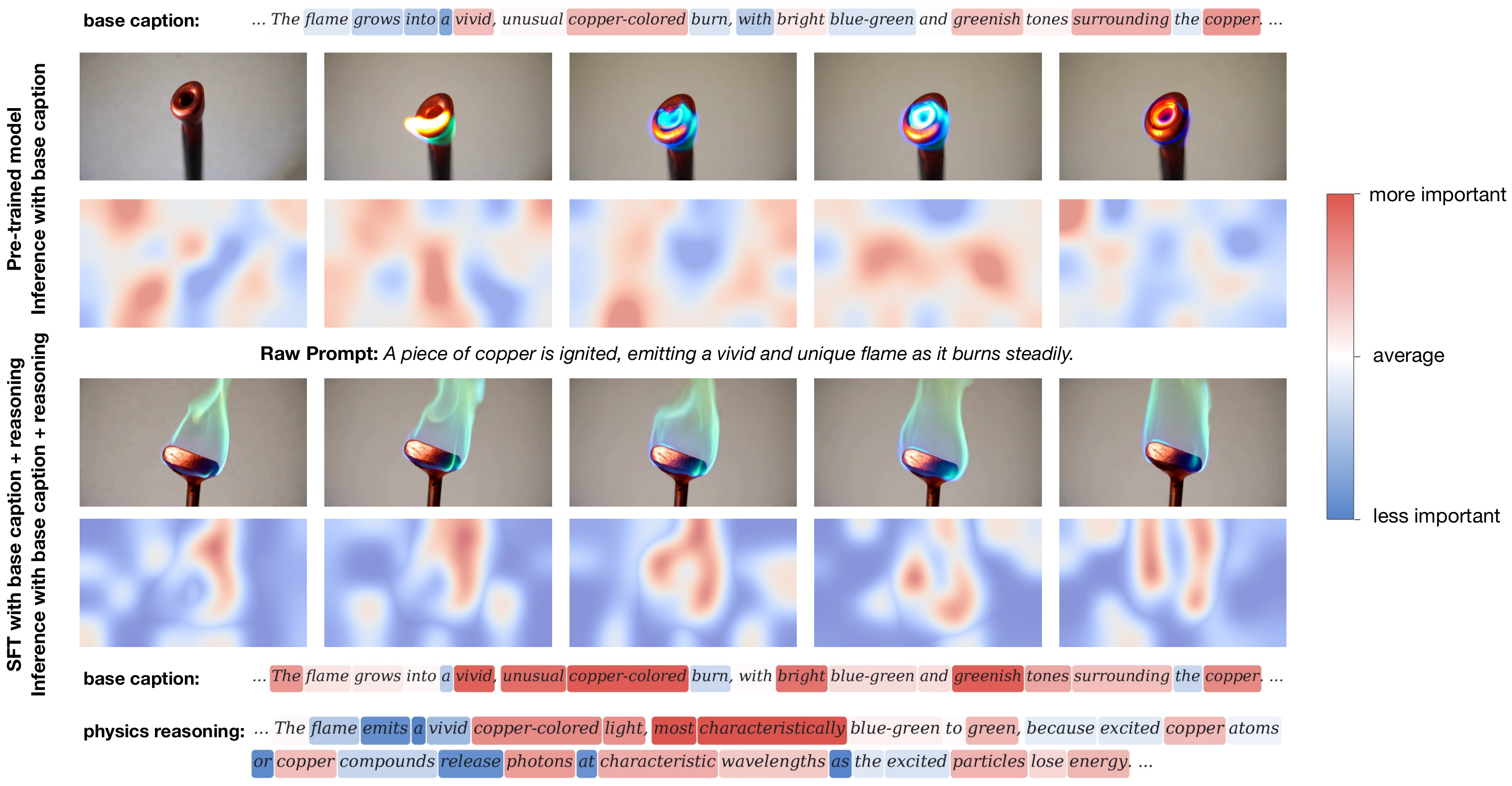}
    \caption{\textbf{Visualization for the attention scores of visual and text tokens.} When adding \texttt{physics\_reasoning} into the SFT and inference process, the generative model can better capture the keywords for the core physical process, and generate videos with physical realism. 
    }
    \label{fig:attention_map}
\end{figure}

As shown in Fig.~\ref{fig:attention_map}, when incorporating \textit{physics\_reasoning} into the SFT and inference process, the model can better locate the keywords which determine the physical process, like ``copper-colored" and ``greenish" in the base caption, and ``photons", ``wavelengths" and ``energy" in the physics reasoning. For the RGB frames, the region depicting the flame color reaction is also more prominent on the attention map.

Moreover, to examine the important role of our fine-tuning in enabling the model to effectively utilize \textit{physics\_reasoning}, we separately feed base caption + \textit{physics\_reasoning} into the pretrained model and the fine-tuned model. Tab.~\ref{tab:attention_map} lists the relative increase in the attention scores of keywords in \textit{physics\_reasoning} after training.

\begin{table*}[h]

\vspace{-6pt}
\centering

\caption{\textbf{Relative increase of attention scores of physical keywords after fine-tuning.}}
\label{tab:attention_map}

\vspace{-6pt}

\resizebox{\linewidth}{!}{
\begin{tabular}{lccc}
\toprule
Caption & $\Delta$ (\%) \\
\midrule
\textit{...Regions directly under fingertips \textbf{COMPRESS} first and most strongly. ...} & 19.2 \\
\textit{...\textbf{HEAT TRANSFER} lowers the average molecular kinetic energy of the juice, ...} & 52.2 \\
\textit{...\textbf{GRAVITY} still acts downward on all gas and particle mass, ...} & 22.5 \\
\textit{...The change in light speed at the lens surfaces \textbf{BENDS} the rays. ...} & 48.7 \\
\bottomrule
\end{tabular}
}
\end{table*}

\clearpage

\section{Validation of GPT-5.5 as a Better Evaluator}
\label{sec:gpt55}

In Sec.~\ref{sec:experimental_details}, we replace the offline VLM evaluators for PhyGenBench and VideoPhy-2 with GPT-5.5. In this section, we illustrate the rationale for this replacement using VideoPhy-2 as an example.

\begin{figure}[h]
    \centering
    \includegraphics[width=\linewidth]{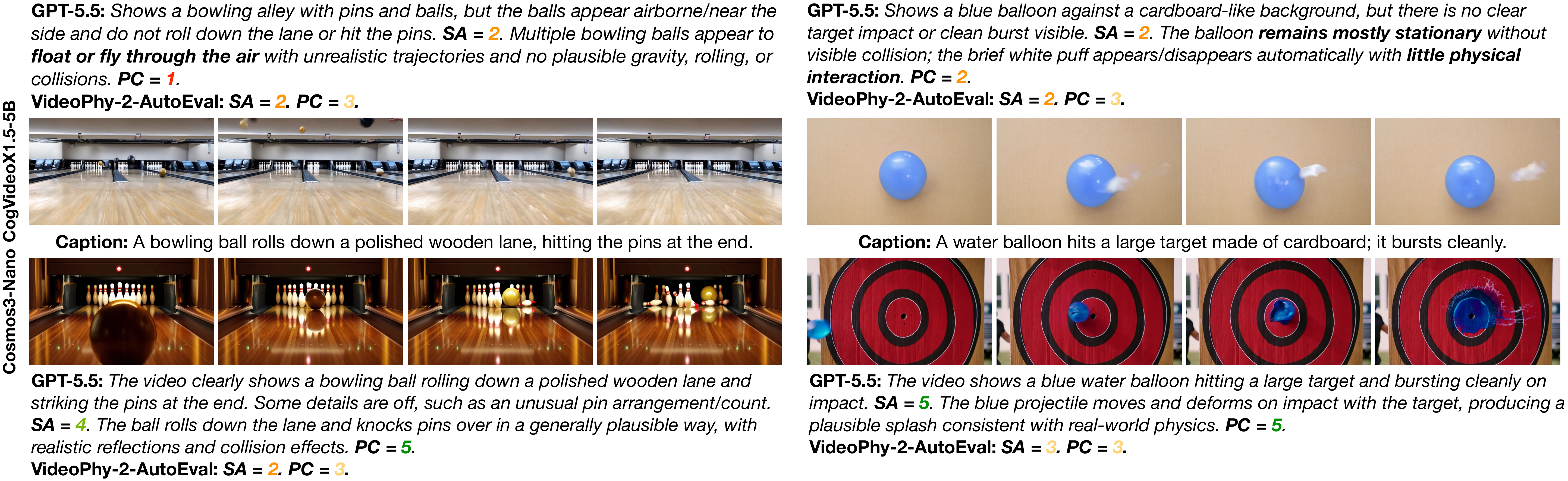}
    \caption{\textbf{Different evaluators for videos generated by CogVideoX1.5-5B and Cosmos3-Nano.} While the auto-evaluator of VideoPhy-2 tends to assign similar mid-range scores to videos with substantially different levels of physical realism, GPT-5.5 provides more discriminative assessments, better reflecting the differences in physical realism across videos.
    }
    \label{fig:gpt55}
\end{figure}

As shown in Fig.~\ref{fig:gpt55} and Tab.~\ref{tab:evaluators}, distinct video generative models often receive nearly identical scores from the pretrained VLM, leading to results that do not faithfully reflect their physical performance. In contrast, GPT-5.5 produces assessments that align with human judgment in most cases.

\begin{table*}[h]

\vspace{-6pt}
\centering

\caption{\textbf{Comparison of VideoPhy-2 scores under different evaluators.} 
}
\label{tab:evaluators}

\vspace{-6pt}

\resizebox{0.5\linewidth}{!}{
\begin{tabular}{lccc}
\toprule
Generative Model & Pretrained VLM & GPT-5.5 \\
\midrule
CogVideoX1.5-5B & 23.01 & 49.41 \\
Wan2.1-T2V-14B & 23.52 & 57.02 \\
Cosmos3-Nano & 22.50 & 60.41 \\
Veo-3.1 & 23.69 & 68.87 \\
\bottomrule
\end{tabular}
}
\end{table*}

\clearpage

\section{General Generative Capability of \methodname}
\label{sec:vbench}

In Sec.~\ref{sec:backbone_scale}, we demonstrate that \methodname{} explicitly optimizes captions and improves the physical realism of generated videos. A natural question is whether such physics-aware captions could compromise the model's general capability of video generation. To investigate further, we evaluate the pretrained and \methodname-enhanced models on VBench for the I2V task. As shown in Tab.~\ref{tab:vbench}, \methodname{} maintains comparable performance to the pretrained models, suggesting that the richer and more structured captions introduced by \methodname{} can provide useful information beyond explicitly physical scenarios.

\begin{table*}[h]

\vspace{-6pt}
\centering

\caption{\textbf{Comparison of different models on VBench.} 
}
\label{tab:vbench}

\vspace{-6pt}

\small
\renewcommand{\arraystretch}{1}
\setlength{\tabcolsep}{7pt}
\begin{tabular}{lccc}
\toprule
\textbf{Model} &
\textbf{VBench-I2V}  \\






\midrule

Wan2.1-14B (Pretrained)  & 86.86 \\

Wan2.1-14B (\methodname)  & 87.51  \\

\midrule

Cosmos3-Nano (Pretrained)  & 88.32  \\

Cosmos3-Nano (\methodname)  & 88.69  \\

\bottomrule
\end{tabular}
\end{table*}

\clearpage

\end{document}